\documentclass{egpubl}
\usepackage[T1]{fontenc}
\usepackage{dfadobe}

\biberVersion
\BibtexOrBiblatex
\usepackage[backend=biber,bibstyle=EG,citestyle=alphabetic,backref=true]{biblatex} 
\electronicVersion
\PrintedOrElectronic

\ifpdf \usepackage[pdftex]{graphicx} \pdfcompresslevel=9
\else \usepackage[dvips]{graphicx} \fi

\usepackage{egweblnk} 
\usepackage{hyperref}
\usepackage{booktabs, siunitx} 
\usepackage{multirow}
\usepackage[normalem]{ulem}
\useunder{\uline}{\ul}{}

\usepackage{amsmath}
\DeclareMathOperator*{\argmax}{arg\,max}

\usepackage{color}

\title{Semantic Segmentation in Art Paintings}

\author[N. Cohen \& Y. Newman \& A. Shamir]
{\parbox{\textwidth}{\centering Nadav Cohen$^{1}$
        and Yael Newman$^{2}$ 
        and Ariel Shamir$^{3}$
        }
        \\
{\parbox{\textwidth}{\centering 
    $^1$The Hebrew University of Jerusalem
    $^2$Tel-Aviv University
    $^3$Reichman University
      }
}
}

\begin{document}
\teaser{
 \includegraphics[width=\linewidth]{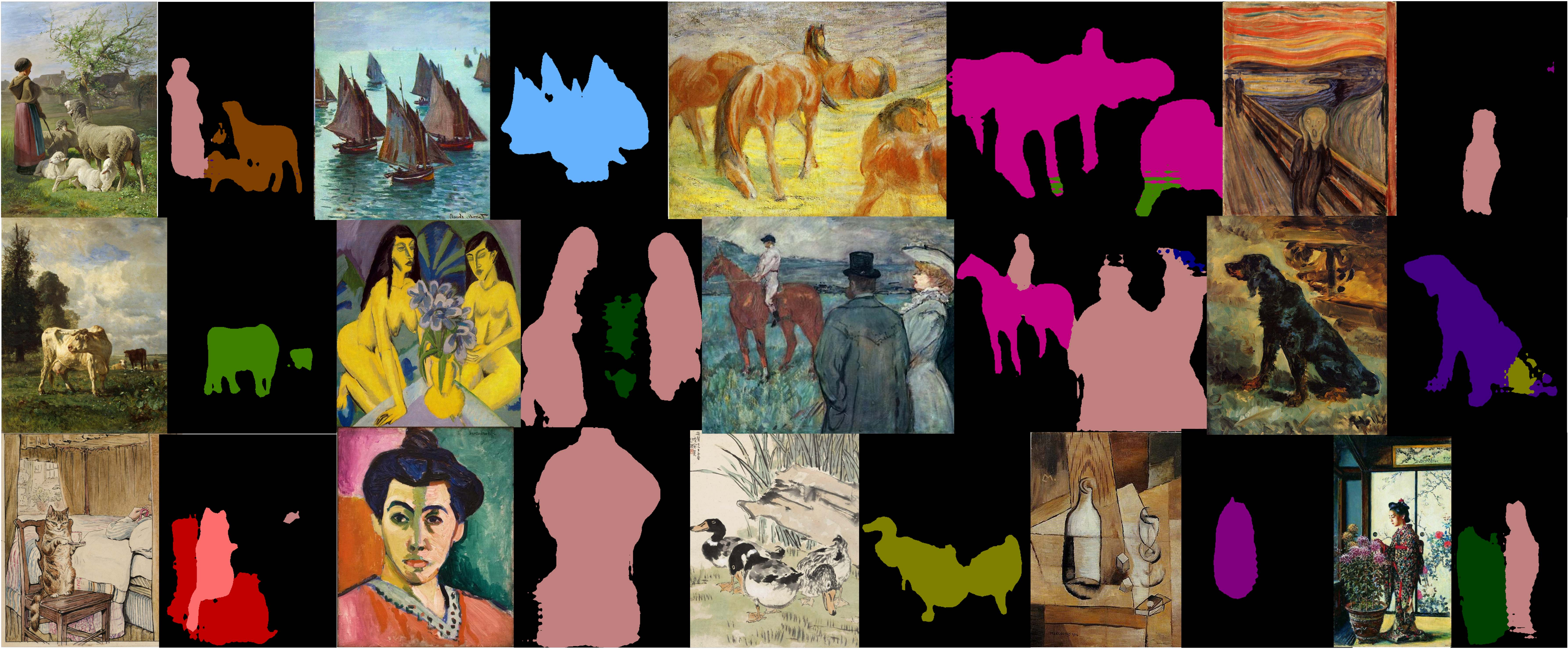}
 \centering
  \caption{Examples of results of our method for semantic segmentation of artistic paintings in various styles -- each color represents a class.  We use unsupervised domain adaptation on the DRAM (Diverse Realism in Art Movements) dataset we collected. DRAM contains figurative paintings from the Realism, Impressionism, Post-Impressionism and Expressionism art movements (the first two rows). The third row shows examples of segmentation of artistic styles which were unseen during training.}
\label{fig:teaser}
}

\maketitle
\begin{abstract}
Semantic segmentation is a difficult task even when trained in a supervised manner on photographs. In this paper, we tackle the problem of semantic segmentation of artistic paintings, an even more challenging task because of a much larger diversity in colors, textures, and shapes and because there are no ground truth annotations available for segmentation. 
We propose an unsupervised method for semantic segmentation of paintings using domain adaptation. Our approach creates a training set of pseudo-paintings in specific artistic styles by using style-transfer on the PASCAL VOC 2012 dataset, and then applies domain confusion between PASCAL VOC 2012 and real paintings. These two steps build on a new dataset we gathered called DRAM (Diverse Realism in Art Movements) composed of figurative art paintings from four movements, which are highly diverse in pattern, color, and geometry.
To segment new paintings, we present a composite multi-domain adaptation method that trains on each sub-domain separately and composes their solutions during inference time. Our method provides better segmentation results not only on the specific artistic movements of DRAM, but also on other, unseen ones. We compare our approach to alternative methods and show applications of semantic segmentation in art paintings. The code and models for our approach are publicly available at: \url{https://github.com/Nadavc220/SemanticSegmentationInArtPaintings}.


\end{abstract}  

\section{Introduction}
\label{sec:intro}
 
\emph{Semantic segmentation} of photographs, where each pixel is assigned to one of a set of predefined classes is a difficult task even using today's methods based on neural networks.
Methods that train with segmented photographic datasets with around 20 classes, such as PASCAL VOC 2012 \cite{pascal} achieve high mean-IOU results  \cite{deepLabv2,deeplabv3p, dis2017, pspNet}. Semantic segmentation becomes even harder in the artistic domain. Artistic paintings have a very different appearance compared to natural photographs,
even when concentrating only on figurative art (i.e.\ non-abstract). They also have much larger diversity in terms of both colors and shapes of objects, and backgrounds.

In this paper, we address the problem of semantic segmentation of (figurative) artistic paintings (see Figure~\ref{fig:teaser}). 
Gathering and annotating an artistic painting dataset is a daunting task, as there are numerous styles and genres within the artistic domain. Hence, our work builds an unsupervised solution using domain adaptation that not only provides a segmentation solution to paintings in some pre-defined artistic styles, but also allows to segment paintings in unseen styles. 

Our method uses two steps. The first step creates a pseudo training-set in some predefined artistic styles (we used Realism, Impressionism, Post-Impressionism and Expressionism) by using style-transfer methods on the existing photographic ground-truth data of PASCAL VOC 2012. We call such datasets \emph{pseudo-paintings} and use them to train basic semantic segmentation networks for these styles. In the second step, we further refine these networks by using a domain confusion technique using PASCAL VOC 2012 as the segmented ground truth source domain and real artistic paintings as the target domain. To segment a new painting, not necessarily from the original domain styles, we first map it to a style latent-space and then combine the segmentations produced by our trained networks based on the similarity of the painting to each domain. 


Previous domain adaptation solutions for semantic segmentation mostly concentrate on adapting synthetic rendered images to real photographs (e.g.\ using GTA5 computer game \cite{gta5} to CityScapes dataset \cite{cityscapes}). Our work requires the opposite direction -- adaptation of real photographs to synthetic, artistic data. We show that simple use of existing domain adaptation techniques does not provide much gain. Even using their original data, reversing the adaptation direction of these methods (i.e. adapting CityScapes to GTA5) reduces the success rate considerably (see supplemental material). This means that adaptation, in general, is \emph{not symmetric} in terms of domains, and there is a need for specialized solutions to adapt photographs to the synthetic domain of paintings. 

We see two main reasons for the difficulty of adapting the segmentation of photographs to paintings. The first reason involves the \emph{domain gap}: there are large differences in the characteristics of artistic paintings compared to real photographs. The second reason is \emph{domain diversity}: artistic paintings cannot be seen as a single coherent domain for learning as they encompass a plethora of styles and movements. Our proposed method tackles both challenges. To tackle the domain gap we use style transfer to create pseudo-paintings for training in the first step. To tackle the domain diversity, we separate the target artistic domain to sub-domains and build a \emph{multi-domain adaptation} solution by combining their results during inference. 

To guide both the style transfer step and the domain confusion step we use a new 
dataset we gathered called DRAM: Diverse Realism in Art Movements. DRAM was intentionally created with high variability and large domain gaps. It is comprised of figurative paintings from four art movements: Realism, Impressionism, Post-Impressionism, and Expressionism. These art movements have highly diverse pattern and geometric styles. The objects and scenes painted do not always appear in their true colors and patterns, and their geometric structure is often distorted (see Figure~\ref{fig:teaser}). We use DRAM as the target data for adaptation.  
For our source dataset we chose PASCAL VOC 2012 \cite{pascal} as it contains a significant number of classes which are more common in classic artwork.

\begin{figure*}[t]
\begin{center}
\includegraphics[width=1.0\textwidth]{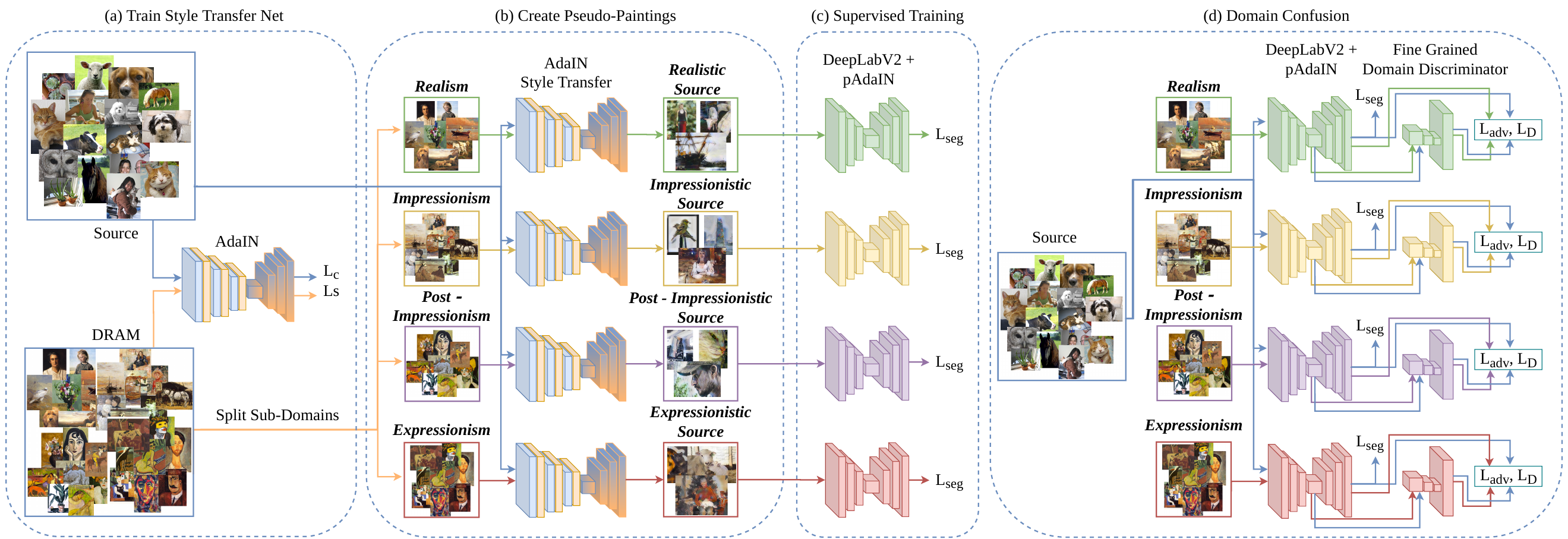}
 \hfill
    \caption{Our proposed training flow: (a) we train a style transfer network using PASCAL VOC 2012 source data as content and DRAM training set as style images; (b) we split DRAM into four sub-domains and create pseudo-paintings for each by augmenting the original source dataset with each sub-domain as style images; (c) we train each of the domains separately in a supervised manner using DeepLabV2 \cite{deepLabv2} on the pseudo-paintings with their original labels from PASCAL VOC 2012; (d) we continue to refine the networks with a domain confusion step based on FADA \cite{fada} and pAdaIN \cite{padain}. We train with PASCAL VOC 2012 as source (without augmentations) and the DRAM sub domains as targets.  } 
    \label{fig:overview}
    \vspace{-0.5cm}
\end{center}
\end{figure*}

Figure~\ref{fig:overview} provides an overview of our approach. We first train a style transfer network on the DRAM data, but use it separately for each sub-domain to create pseudo-paintings of each artistic movement, capturing its unique characteristics. Next, we train semantic segmentation networks using these pseudo-paintings with their original segmentation labels. Lastly, we apply adversarial domain confusion to further refine the segmentation network of each sub-domain using DRAM's real paintings. During inference, given an input painting, we first map it to a style feature-space using Gram matrices \cite{gatys2015} as style descriptors. In this space, we find its k-nearest neighbors from the mapped paintings of the DRAM dataset. Based on the ratio of neighbors belonging to each sub-domain we use a weighted combination of the sub-domains segmentation solutions to segment the input painting (see Figure~\ref{fig:inference}).
Our experiments show that using such multi-domain inference can be used to segment paintings from unseen artistic domains, but surprisingly it also improves the segmentation results of paintings from the original sub-domains.

To summarize our contributions are:

\begin{itemize}
     \item We present the first semantic segmentation solution for artistic paintings.
    \item Our method combines multi-domain adaptation for the highly diverse domain of art paintings. 
   \item We present the DRAM dataset: a new artistic domain adaptation benchmark with a fully segmented test set.  
\end{itemize}

We present results of segmenting artistic paintings, experiments of an ablation study, and show  applications of our method. The new DRAM artistic benchmark as well as our code will be released for future research. 
We see their contribution not only for semantic segmentation of art paintings, but also to domain adaptation development in general, by challenging generalization in highly diverse domains, and testing adaptation from the real domain to a synthetic domain. 

\section{Related Work}
\label{sec:prev}

\textbf{Semantic Segmentation.}
The leading approach for semantic segmentation of images uses CNNs \cite{unet, fastfcn, maskrcnn, deepLabv2, deeplabv3p, hrnet}. Given an input image, a neural network outputs a label per pixel in the image.  Current state-of-the-art results on photographs of different datasets are achieved with DeepLabV3+~\cite{deeplabv3p}. DeepLabV3+ uses a classification network as its encoder and up-samples the encoding output using atrous convolution layers followed by a simple convolutional decoder. We follow current domain adaptation methods and use DeepLabV2~\cite{deepLabv2} as the basic semantic segmentation network. DeepLabV2 does not use a convolutional decoder to up-sample its encodings. We compare our results to both DeepLabV2 and DeepLabV3+ as baselines. As recent papers such as \cite{ocr, swin} have shown promising improvement on semantic segmentation by using modules based on self-attention transformers \cite{attention, vit} we use HRNet \cite{hrnet} with OCR  \cite{ocr} as an additional baseline for comparison. As artistic paintings have differences in color, textures and geometric structure of objects, none of the three methods perform well in this domain when trained on real-life photographs.

A previous attempt to improve segmentation results on artistic paintings was made by Chatzistamatis et al. \cite{color_blind} which focuses on recoloring art paintings to overcome color loss caused by color blindness by  utilizing segmentation maps. To do so a pretrained MaskRCNN \cite{maskrcnn} is fine-tuned using annotated art paintings to output semantic maps which are used for the semantic recoloring process. Unlike Chatzistamatis et al. our work focuses on semantic segmentation of art in unrealistic challenging styles as well as over unseen art styles. Additionally, we present a unique unsupervised solution for semantic segmentation of art paintings and we compare our results to other recent methods.


\textbf{Domain Adaptation (DA).}
Domain adaptation works with two datasets drawn from two different domains: a labeled dataset for the \emph{source} domain and an unlabeled dataset for the \emph{target} domain. The data of the target domain is the one we wish to optimize on a given task.
Initial frameworks of domain adaptation targeted the image classification task and centered around adversarial domain confusion \cite{Ganin2015DA, adda, gvbDA}. Later classification frameworks tackled more advanced challenged like multi-target domain adaptation \cite{blendingtarget, infoApproach} and some used artistic datasets \cite{domainnet, gvbDA, infoApproach}. However, unlike our DRAM dataset the artistic datasets explored by these papers focused on the difference between different type of arts such as clip art and sketches rather than the subtle difference between fine art painting styles.

Initial semantic segmentation DA also used domain confusion~\cite{adaptsegnet} and added an image translation module to reduce the gap between the source and target domains~\cite{cycada}. Later solutions mainly rely on three techniques: data augmentation, domain confusion, and self-learning. Each method uses these three techniques differently and may use only a subset of them. In the following, we elaborate on the first two techniques, data augmentation and domain confusion, as we do not use self-learning in our method. Additionally, we discuss common DA datasets and prior work related to our dataset challenges: target domain diversity and a large domain gap.

\textbf{Datasets.}
As research in artistic domains is mainly focused around practical applications such as style transfer and art creation \cite{gatys2015, adain-style-transfer, faceOfArt} most artistic datasets \cite{wikiart,WikiPaintings} do not come annotated for object identification or semantic segmentation. Other, more perceptual applications include artist/genre clustering~\cite{selective_cluster} and art paintings classification \cite{WikiPaintings}. Exceptions to this are \cite{art_oxford, domainnet, pacs}, but they include annotations only for image classification and not for semantic segmentation.

Current domain adaptation approaches focus on driving datasets where the source domain is 3D computer renderings and the target domain is realistic photographs. The most commonly used are GTA5 \cite{gta5} or Synthia \cite{synthia} as the (synthetic) source domain, and Cityscapes \cite{cityscapes}, BDD100K \cite{bdd100k}, or Cross-City \cite{cross-city-paper} as the (realistic) target domain. 
In our paper we focus on the opposite transition, from realistic to synthetic domains, using artistic paintings as our target data. We created the DRAM dataset which focuses on semantic segmentation and presents a new and more general benchmark for the unsupervised domain adaptation task. 

\textbf{Data Augmentation.} 
Augmentation is used to reduce the domain gap between the source data and the target data. Beyond simple geometric transformations, different image-level augmentations include CycleGAN \cite{bdl, pceda, cycada}, Fourier-Domain window exchange \cite{fda} and Style Transfer as used by Banar et al. \cite{musical_instruments} to classify musical instruments in art paintings. Nuriel et al. \cite{padain} utilize an AdaIN layer \cite{adain-style-transfer} randomly on different latent features to exchange information between the target and source data to reduce the pattern bias shown in \cite{texture-bias} and \cite{gatys2015}. Li et al. \cite{bdl} and Yang el al. \cite{pceda} also optimize the domain transformer between the framework learning steps to improve transformations by using information from the learning process.

Using CycleGAN~\cite{imgtrnsltcycleGAN} as a transformation network from PASCAL VOC 2012 to DRAM resulted in unsatisfying results. We believe this may be due to the high complexity and diversity of the artistic domain. The DRAM training set contains over 50 different artists and we suspect that optimizing a single network for such a diverse domain may be too complex.
We use AdaIN style transfer~\cite{adain-style-transfer} for augmentation to reduce the domain gap between the realistic photographic source data and our artistic target data, by creating \emph{pseudo-paintings}. We chose this method for its ability to preserve the overall look of the source image and for its speed, which enables augmenting large datasets relatively fast.

\textbf{Domain Confusion.} Domain Confusion is an adversarial method which utilizes a domain discriminator. The discriminator is used to train the semantic segmentation network's encoder to encode target and source samples to a joint domain, while optimizing the network over the labeled source data in a supervised fashion.
In an effort to use the segmentation information extracted from the network, Tsai et al.~\cite{adaptsegnet} uses two discriminators, one for encoder features and the other for output features. We follow FADA~\cite{fada} and pAdaIn~\cite{padain} which use a discriminator not only to distinguish between domains, but also to learn the class structure of the trained model to encourage a class-level alignment in the generated feature space.


\textbf{Target Domain Diversity.} Most domain adaptation methods assume that the target data is homogeneous. The setting where there are several different sub-domains comprising the target domain is known as Multi-Target Domain Adaptation \cite{openCompoundDA, cross-city-paper, ocda2, blendingtarget, infoApproach, collabrative}. In some, even the target domain labels are unknown \cite{openCompoundDA, ocda2, blendingtarget}. We used painting labels as many datasets classify artistic paintings to their style based on expert knowledge \cite{wikiart}. In addition, we found that utilizing clustering of paintings in some feature space led to inferior segmentation results. 

For Multi-Target Domain Adaptation, Liu et al.~\cite{openCompoundDA} use a curriculum training procedure where target images are used for training in order of their distance from the source domain. Park et al.~\cite{ocda2} train a single segmentation network, but use a separate discriminator for each sub domain to separate their optimization process. Additionally, they perform a unique image transformation process for each sub-domain to better use the assumed separation of the sub domains. Isobe et al.~\cite{collabrative} trains an ``expert'' segmentation network using AdaptSegNet~\cite{adaptsegnet} for each sub domain and then trains another network using DA which uses the information learned by the expert networks.

As discussed in \cite{collabrative}, a solution for a multi-target domain adaptation task, which on the one hand creates a single framework for flexible predictions of each sub-domain, and on the other hand achieves results that are as optimal as training with each domain separately, creates a challenging trade off. 
Our approach trains a separate segmentation network for each sub domain. Similarly to Park et al.~\cite{ocda2} we train a unique image transformation network for each sub domain and apply it on our source data to reduce the domain gap. To create a flexible multi-domain framework we compose the networks predictions by the similarity of the input image style to the training images style, using the style representation presented by Gatys et al.~\cite{gatys2015}.




\textbf{Domains Gap} Domain Adaptation methods rely on a reasonable gap between the source and target domains to achieve good results. When the target domain is incoherent, reducing this gap becomes a nontrivial task as each target sub-domain may require a different approach. Current methods discussing this issue, such as  \cite{cross-city-paper,openCompoundDA}, focus on class-level alignment, which helps align diverse sub-domains by class information rather than relying solely on global domain information. Dai et al.~\cite{domainbridges} creates new ``bridging'' target domains which are trained on separate discriminators and help the network optimize as they are closer to the source domain, thus easing the large gap between the source and the target domain. Our method also utilizes class information in the domain confusion step. Furthermore, we show that for complex domains, understanding the unique properties of each sub-domain can help reduce the domain gap. As a result, we can achieve better results for each sub-domain separately and for the entire target dataset together, as well as for unseen artistic target domains.   


\begin{figure*}[t]
\begin{center}
\includegraphics[width=1.0\textwidth]{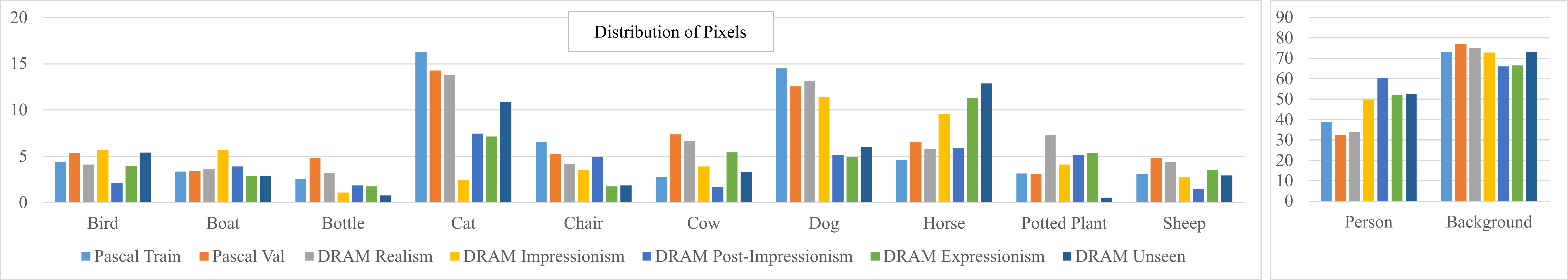}
    \caption{The distribution of number of pixels per each class in PASCAL VOC 2012 training and validation sets compared to all movements in the DRAM test set. For the background class, we show the percentage of pixels labeled "Background" from all pixels. For all other classes, we present the percentage of pixels labeled per class excluding the background class to make the statistics more visible. Perfect equalization is difficult as some classes are more scarce in art than in real life photos.
 %
}
    \label{fig:set_stats}
    \vspace{-0.5cm}
\end{center}
\end{figure*}

\section{DRAM Dataset}
\label{sec:dataset}

To our knowledge, the Diverse Realism in Art Movements (DRAM) dataset, is the first semantic segmentation dataset which uses artistic paintings as its target domain. The dataset is composed of 5677 unsegmented training images and 718 segmented test images of paintings of 152 different artists. The majority of the dataset (including all training images and 583 test images) is comprised of four diverse art movements: Realism, Impressionism, Post-Impressionism, and Expressionism.

The DRAM dataset was gathered mainly from the WikiArt art database \cite{wikiart} (5677 train images, 676 test images). The remaining images (42 test images) were gathered from Wikimedia \cite{wikimedia} and Wikioo \cite{wikoo} image databases. The images were assigned to art movements using their original tags. To ensure we are learning an artistic movement rather than the style of a specific artist, no artist is shared between the training and the test sets of each movement. 


\subsection{Domain Diversity}

Four art movements were chosen for DRAM as they differ in terms of texture, geometry, and color.  Their respective sizes are 1074, 1538, 1462 and 1603 images for training and 150, and 150, 142, 141 images for testing.

\textbf{Realism}: Emerged in France around the 1848 revolution. It sought to portray people from all classes of society in everyday situations with as much truth and accuracy as possible, without avoiding unpleasant aspects of life. For our purpose, this dataset is considered realistic in texture, geometry, and color; we consider it as the closest domain to our source dataset. To expand the Realism training set we added images from the Romanticism art movement as it shares realistic aspects with Realism in spite of it having more dramatic motifs.

\textbf{Impressionism}: Emerged in France in the 19th century. Characterized by thin yet visible brush strokes and an accurate depiction of light. This art movement is highly diverse in texture but is mostly realistic in geometry and color.

\textbf{Post-Impressionism}: Emerged in France roughly between 1886 and 1905 as a reaction against Impressionists' concern for realistic depiction of light and colors. Post-Impressionism emphasizes more abstract qualities and symbolic content. It has mild texture diversity, and is more diverse in geometry and colors.

\textbf{Expressionism}: Originated in Northern Europe around the beginning of the 20th century. Expressionist artists sought to express the meaning of emotional experience rather that physical reality. Expressionist paintings are highly diverse, using radically distorted geometry that represents a strong emotional effect. Expressionism does not oblige to any realistic concepts and is highly diverse in geometry and colors. It can also be diverse in texture, but not as strong as Impressionism. 

The remaining 135 test images were gathered from eight art movements which do not appear in the training set: Art-Nouveau, Baroque, Cubism, Divisionism, Fauvism, Chinese Ink and Wash, Japonism and Rococo. As the training set does not include examples from these art movements, we consider these test images as unseen data to demonstrate the ability of our method to predict segmentation maps on data from unseen art movements that possibly present a larger domain gap and diversity.

Another aspect causing diversity in art movements is the choice of motifs. For example, the Baroque and Rococo movements are both fairly realistic styles, but Baroque style depicts elements from the Catholic Church with a strong religious atmosphere, while Rococo depicts reality in a more theatrical sense. Both movements appear in our unseen test set for validating generalization ability on such artistic variations.


\subsection{Domains Gap}

We  use PASCAL VOC 2012~\cite{pascal} as our source dataset with 11 classes common in art: Bird, Boat, Bottle, Cat, Chair, Cow, Dog, Horse, Sheep, Person, and Potted-Plant. The rest of the classes and any other object not recognized in our task are considered background, which is used as the 12th class in our tests.
The test data in DRAM was manually segmented according to PASCAL VOC 2012 official guidelines with the only exception being labeling flower vases as potted-plant to expand the number of classes available (they share similar features and potted plants are scarce in most art movements). 

To reduce the domain gap in terms of content and ensure a fair evaluation of our results, we took effort to equalize the class statistics between DRAM and PASCAL VOC 2012 (see Figure \ref{fig:set_stats}). 
Otherwise, any statistical measure could have been biased because of lack of a specific class or domination of another. We used an iterative process of adding paintings to the DRAM dataset while preserving similar distribution of classes. 

\begin{figure*}[t]
\includegraphics[width=0.5\columnwidth]{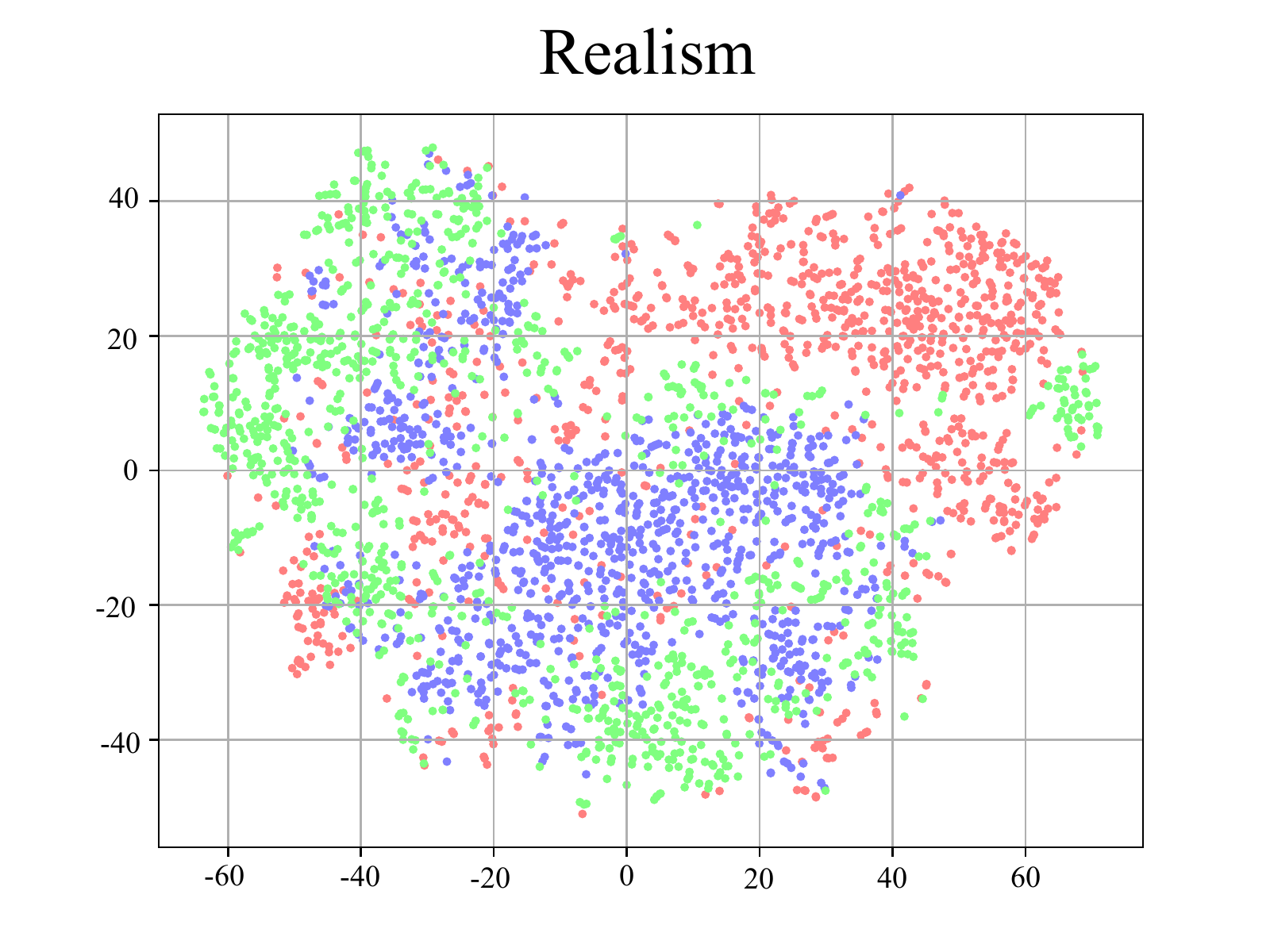}
 \hfill
\includegraphics[width=0.5\columnwidth]{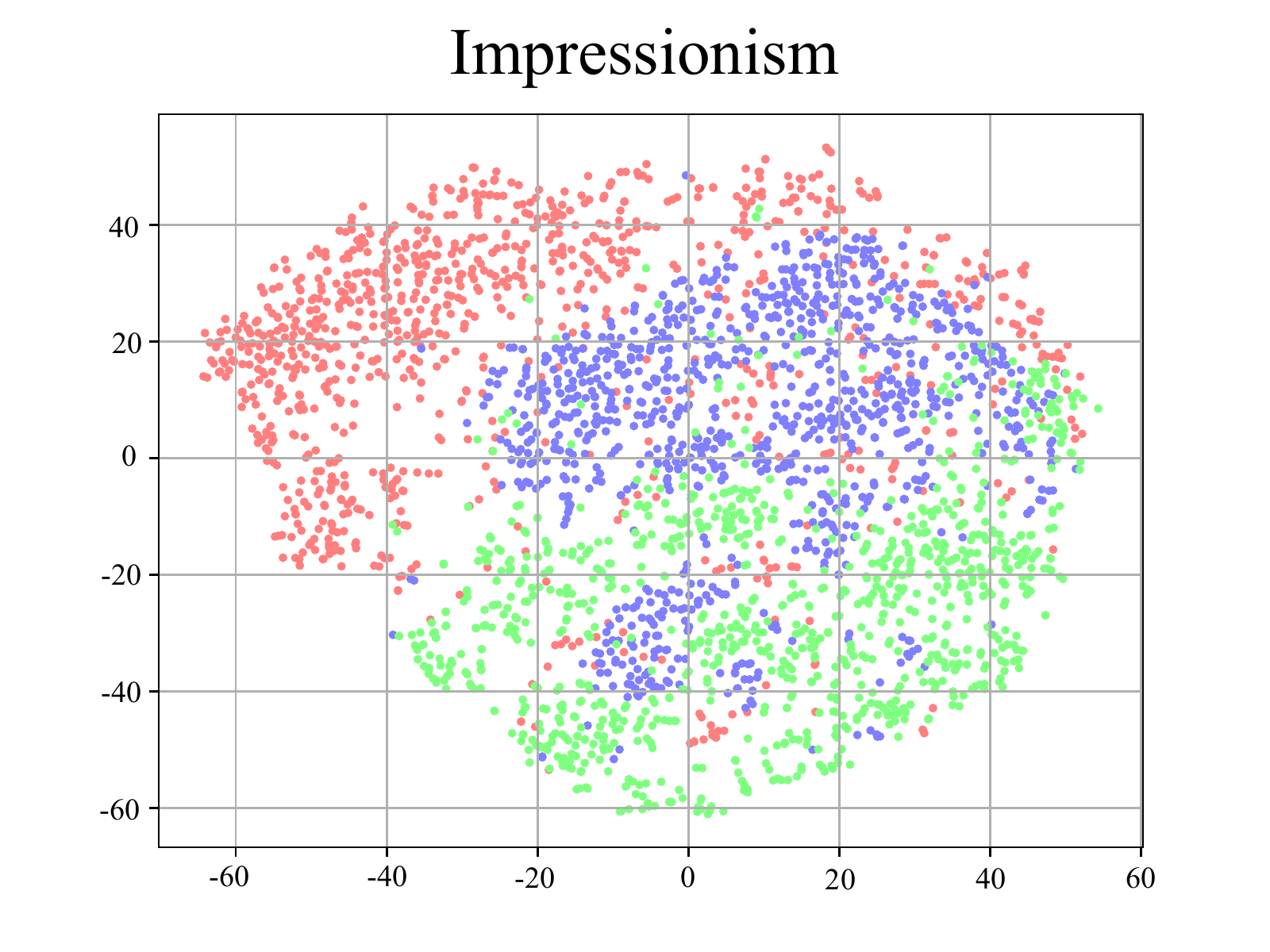}
 \hfill
\includegraphics[width=0.5\columnwidth]{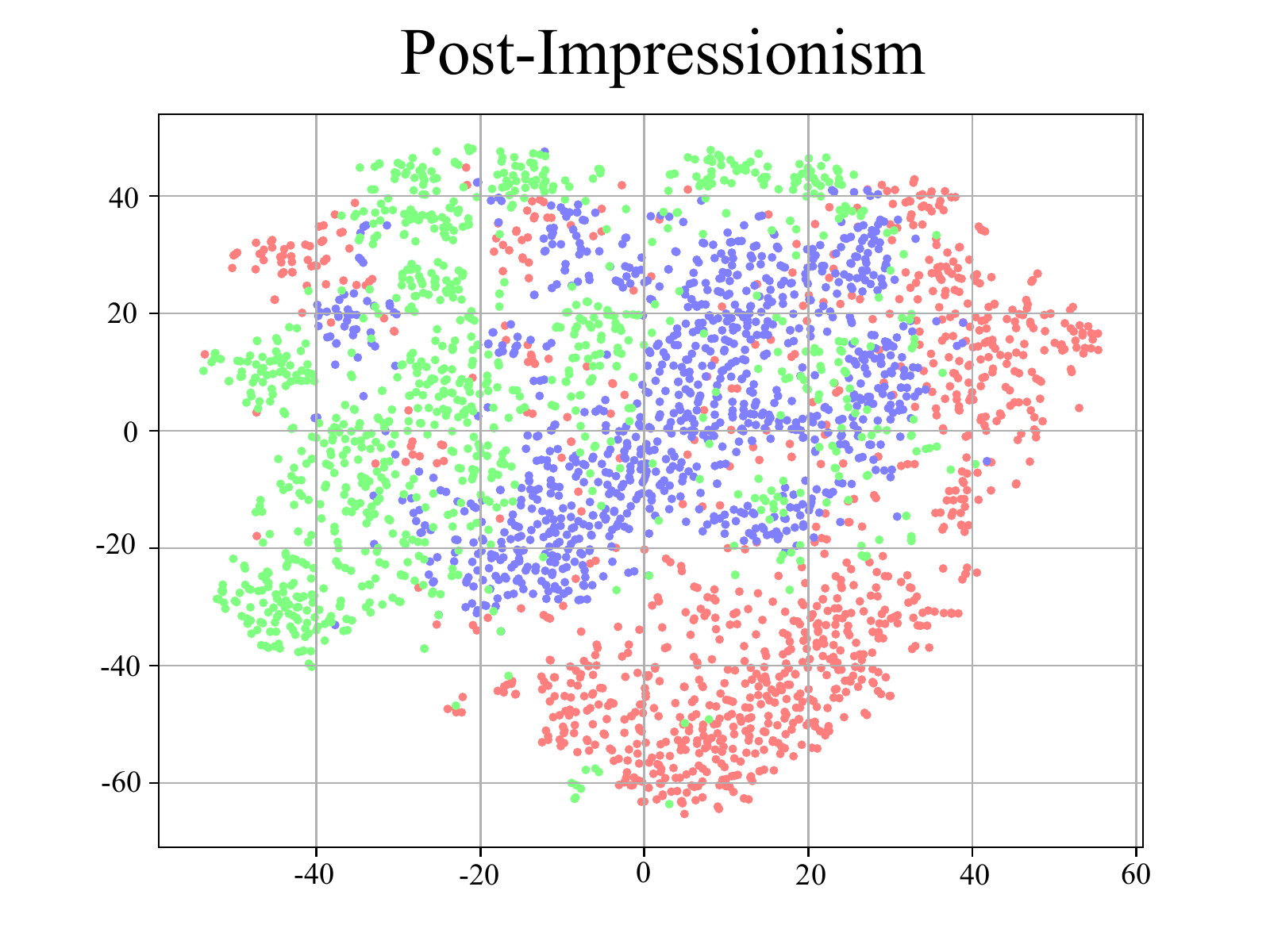}
 \hfill
\includegraphics[width=0.5\columnwidth]{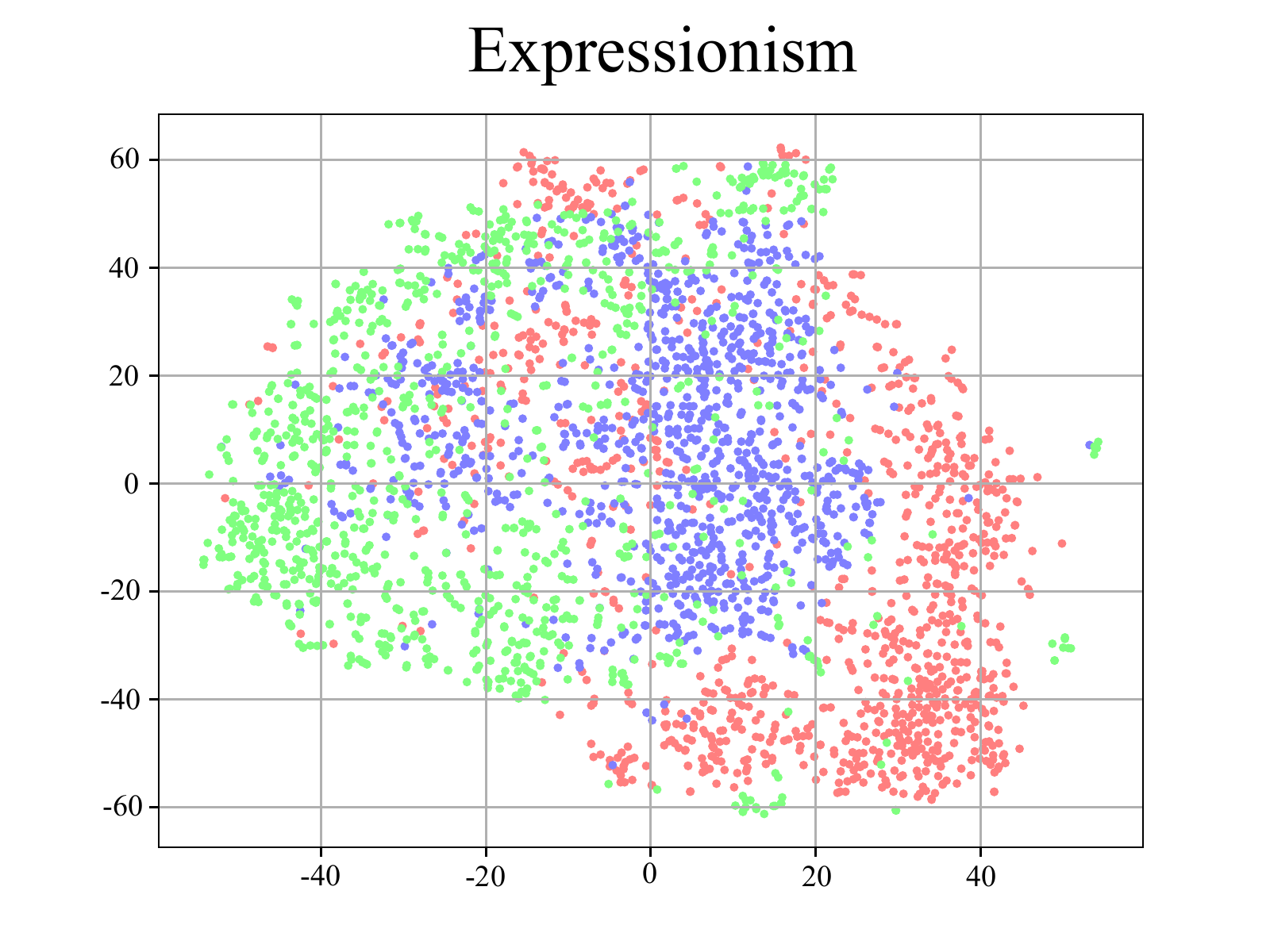}
 \hfill
    \caption{TSNE plots of the style feature space of the four art movements in the DRAM dataset (green dots) compared to PASCAL VOC 2012 dataset (red dots) -- the gap between each synthetic domain and the real domain is clear. From left to right we plot Realism, Impressionism, Post-Impressionism, and Expressionism. We use style-transfer to create a set of pseudo-paintings in each movement (blue dots) that bridge the gap between the domains, which are used for training in the first step of our method. }
    \label{fig:tsne}
    \vspace{-0.5cm}
\end{figure*}

\subsection{Style Feature Space}
The seminal work of Gatys et al.~\cite{gatys2015} introduced the concept of Gram matrices, which are obtained by matrix multiplication of VGG19 convolution layers, and serve as an expression of the artistic texture-style of a given painting. 
We use this representation as a style feature space for our data. 
We concatenate the Gram representation of 5 pre-trained VGG19 layers: $conv11,  conv21, conv31, conv41$ and $conv51$ as suggested by \cite{gatys2015}. For efficiency purposes, we use Kernal-PCA~\cite{KernalPCA} with a cosine kernel to reduce the representation dimensions to 512, preserving more than 99\% of the data variability.
Figure~\ref{fig:tsne} shows a TSNE plot~\cite{tsne} of the mapping of paintings from DRAM's four art movements compared to the mapping of PASCAL VOC 2012's photographs. The gap between the domains in terms of style is clear. To reduce this domain gap we use style transfer in the first step of our method as described in the next section.




\section{Method}
\label{sec:method}
Given a target domain $X_t$, an image sampled from the target domain $x_t\in{X_t}$ and a finite set of classes $C=\{c_1, ... c_{|C|}\}$ we wish to assign the correct class to every pixel in $x_t$ using the output of a chosen semantic segmentation network:
\begin{equation}\label{semantic prediction eq}
P_{x, y} =\argmax_ {c_i\in C} \phi_{x,y}(x_t)
\end{equation}
where $\phi(\cdot)$ is the semantic segmentation model trained with domain adaptation and $\phi_{x,y}(x_t)$ is the vector of size $|C|$ taken from the $(x, y)$ pixel location of the segmentation model output  $P = \phi(x_t)$.

To train semantic segmentation on artistic paintings without excessive tagging we turn to an unsupervised domain adaptation framework, where the source domain (PASCAL VOC 2012) includes real photographs with ground-truth segmentation, and the target domain (DRAM) contains unsegmented paintings.

Most current domain adaptation methods also consider the target dataset samples as drawn from a single coherent domain. However, some image domains $X_t$ may contain more than one sub-domain, and training it as a single domain achieves sub-optimal results due to the sub-domain differences. In our case, art paintings can differ considerably by many factors such as texture, color, and geometry. Treating many art movements as a single image domain produces a highly diverse domain which hurts the adaptation results.

To compensate for the large diversity, we suggest training each art movement as a single coherent sub-domain (see Figure~\ref{fig:overview}). Doing so enables the trained sub-models to learn specific features relevant only to the specific art movement and prevents irrelevant features from misguiding the optimization. During inference, we combine these sub-models to segment not only paintings from the original domains, but also paintings from other, unseen art movements.

\begin{figure*}[t]
\begin{center}
\includegraphics[width=1.0\textwidth]{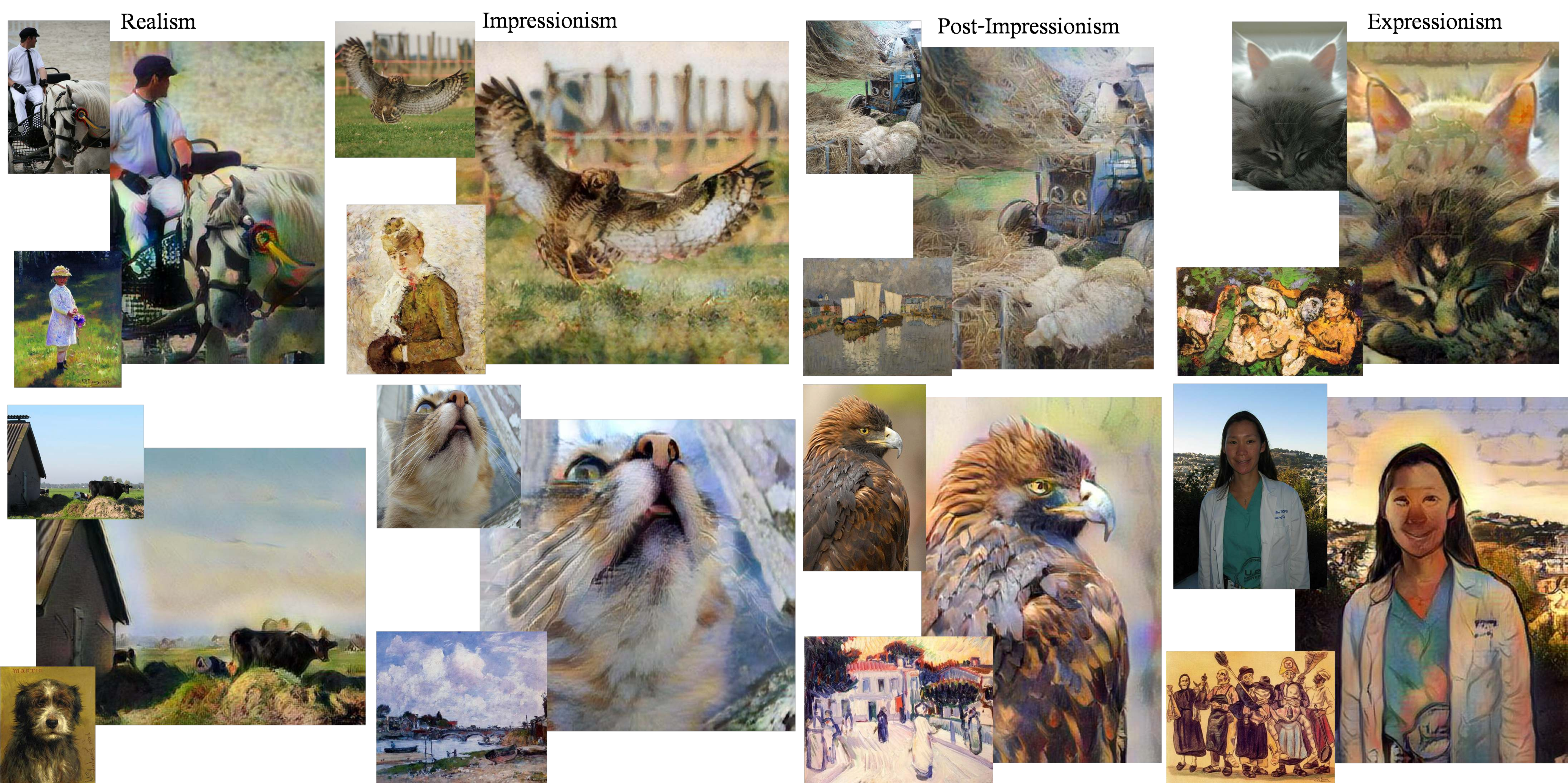}
    \caption{
    Examples of pseudo-painting for the four art movements in the DRAM dataset from left to right: Realism, Impressionism, Post-Impressionism, and Expressionism. Each example shows the original photo and the original painting (left) used to create the pseudo-painting (right).}

    \label{fig:pseudo_paintings}
    \vspace{-0.5cm}
\end{center}
\end{figure*}

\subsection{Augmentation and Pseudo-Paintings}
\label{methodst}

The first step of our method utilizes style-transfer to create a dataset of segmented pseudo-paintings to use as a training set for a segmentation network. This step bridges the gap between the source and the target domain for training.

We first train a style transfer network using the style transfer approach suggested by Huang et al.~\cite{adain-style-transfer}. This approach introduced the AdaIN layer that swaps statistics at the feature level from a style image to a content image we wish to stylize (Figure~\ref{fig:overview} (a)). One advantage of this method is that it can be trained on a collection of style images and then applied on arbitrary style and content images. We use the entire DRAM training set as style images and PASCAL VOC 2012 as content images (more details can be found in the supplemental materials).

Rather then using a single augmented dataset as common in current domain adaptation approaches, we augment the source dataset separately for each sub-domain. We use the original classification of the artwork and create four sub-domains in DRAM, namely: Realism, Impressionism, Post-Impress\-ionism, and Expressionism. Each sub-domain is used separately as style images for augmenting the source domain and create a separate set of pseudo-paintings in the four different artistic styles (Figure~\ref{fig:overview} (b)). Figure~\ref{fig:tsne} clearly shows how the new pseudo-painting datasets are closer in style feature space to the original paintings of each movement, and how the pseudo-paintings bridge the gap between the photographic domain of PASCAL VOC 2012 and the actual paintings. Examples of pseudo-paintings used for training our segmentation networks can be observed in Figure \ref{fig:pseudo_paintings}.



Using the pseudo-paintings of each sub-domain we train a semantic segmentation network for each sub-domain in a supervised fashion (Figure~\ref{fig:overview} (c)).
As can be observed in Table~\ref{table: comparison}, this training with augmented pseudo-paintings alone (marked as \textbf{OurMethod$\setminus$DC}) improves the results of segmentation significantly for all sub-domain compared to the baseline without adaptation (\textbf{DeepLabV2}, \textbf{DeepLabV3+} and \textbf{HRNet+OCR}). This indicates that our style transfer augmentation helps reduce the domain gap between the source domain and each target sub-domain. However, we further adapt the segmentation networks using an additional domain confusion step as described next.



\subsection{Domain Confusion}
\label{methoddc}

We chose to base our domain confusion step on FADA~\cite{fada} with pAdaIN~\cite{padain}, since it achieves better regularization when trained on the source domain. This regularization is an important feature in the artistic domain because of its high diversity.

FADA is a domain adaptation framework which uses three training steps: supervised source training, domain confusion, and self-supervised learning. The domain confusion step uses a fine-grained domain discriminator which receives, in addition to the domain label, soft labels generated by the current network predictions. Doing so, the discriminator learns class information and achieves a better alignment between classes on the encoded latent domain. As discussed in section \ref{sec:prev}, this approach was found to be useful for highly diverse domains.  

pAdaIN uses the framework proposed by FADA and presents a regularization term which reduces texture bias as presented by Geirhos et al.~\cite{texture-bias}. The method suggests adding an AdaIN layer \cite{adain-style-transfer} after every convolution layer in the ResNet encoder to swap image statistics between images among a batch with random probability of 0.01. The AdaIN layer swaps the image statistics while leaving global characteristics such as color and overall structure intact. Note that in the first and third steps of FADA, pAdaIN is used between random images inside a batch, where half of the batch is used as style images which transfer their statistics to the other half of the batch. In the second step, the target data batch is used as style images to transfer its statistics to the source data in the same fashion. More details regarding the training and networks can be found in the supplemental materials.

We have found that using self-supervised learning based on pseudo-labels (which is the third step of FADA) does not always improve the segmentation results on paintings. We believe this is due to the large diversity and gap between the photographic and artistic domains. Instead, our approach turns to combine the individual solutions of each sub-domain both for the known domains of our DRAM set, as well as for new artistic domains, as described next. 



\subsection{Multi-Domain Inference}


Training each sub-domain separately introduces new challenges during inference. Art movements are based on abstract concepts rather than exact rules. It may be difficult to choose the correct sub-domain for every artistic painting, and especially for new ones from movements unseen during training. In addition, images tagged in a certain art movement can be closer to a different movement. 
Images may include artistic features from more than one art movement and can benefit from using the learned models of several art movements.
Therefore instead of applying each sub-model separately, we combine them together using a method we term \emph{multi-domain inference}.

As mentioned before, we use Gram matrices to represent images in style feature space. We pre-compute the Gram-representation of all training data, map them to the $512$-dimensions feature space and store them. 
During inference, for each query image $z$, we search for its k-nearest neighbors in the style feature space and use the ratio of each sub domain as the weight for predictions of the different sub-domains.
We define the weight $w_z^i$ of each sub-domain $i$ in the inference process as the percent of representatives from this sub-domain in the k-nearest neighbors of $z$:
\begin{equation}\label{Gram weight vec}
w_z^i =\frac{1} {k} \sum_{i=1}^{k} 1_i(\theta_z^i)
\end{equation}
where $1_i(\cdot)$ is the indicator function for the i'th sub-domain, and $\theta_z$ is the set of k-nearest neighbors of $z$.




\begin{figure}[t]
\begin{center}
\includegraphics[width=1.0\columnwidth]{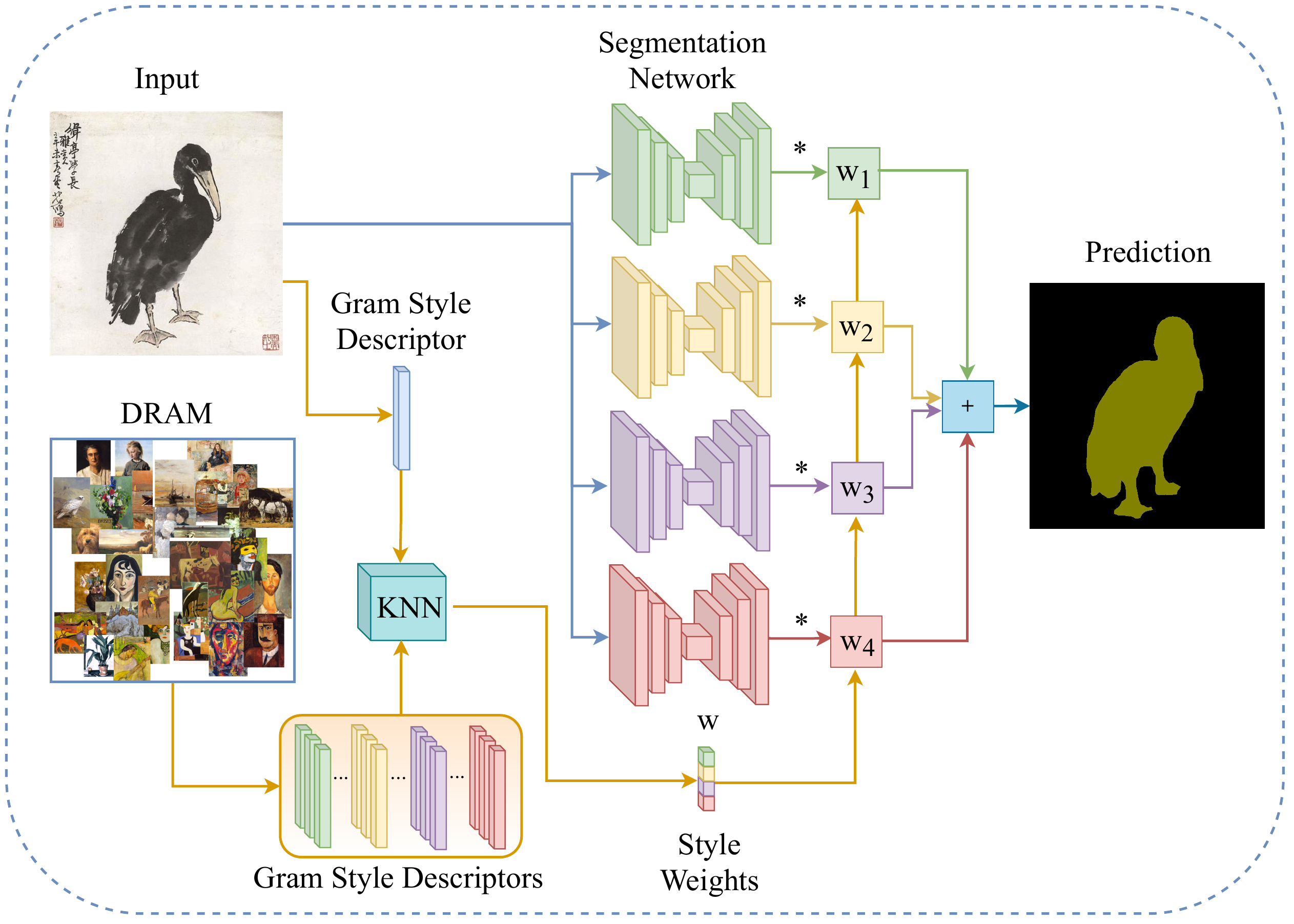}
 \hfill
    \caption{Multi-Domain Inference pipeline. The input image is processed by each pre-trained art movement segmentation network. Then, a weight vector is calculated based on similarities of the image to the training set in Gram feature space and is used to combine the networks outputs and create a final segmentation prediction.  }
    \label{fig:inference}
    \vspace{-0.5cm}
\end{center}
\end{figure}

If $\{\phi^{i}(\cdot)\}_{i=1}^n$ are the semantic segmentation models trained on the $n$ sub-domains of $X_t$ using domain adaptation, we use the weight vector $w_z$ to combine the outputs of $\{\phi^{i}(\cdot)\}_{i=1}^n$ (see Figure~\ref{fig:inference}). Thus, the flexible prediction of an unseen image $z$ is defined for each pixel $(x,y)$ as: 
\begin{equation}\label{equation: weighted pred eq}
Q_{x, y} =\argmax_{c\in C} \sum_{i=1}^{k}w_z^i \cdot \phi_{x,y}^i(z) 
\end{equation}

We use $k=500$ for the KNN search. This $k$ value allows greater precision, while smaller values tend to be more sensitive to outliers (e.g.\ using $k=10$ results in a loss of up to 0.3\% in accuracy).

Our experiments show that multi-domain inference can benefit results not only for new unseen art movements, but also for images of the original art movement sub-domains of the training set. This indicates that paintings from one art movement can contain features from more than one art movement.

\section{Experiments}
\label{sec:experiments}

\begin{table*}[t]

\begin{center}
\begin{tabular}{l| ccccc |c}
\toprule
Method & Realism & Impressionism & Post-Impressionism & Expressionism & Unseen & DRAM \\
\midrule
DeepLabV2 & 52.45 & 36.26 & 30.57 & 15.22 & 34.31 & 34.01 \\

DeepLabV3+ & 60.02 & 39.27 & 29.40 & 15.49 &  30.59 & 35.17 \\

HRNet+OCR & 52.47 & 36.50 & 40.77 & 20.01 &  34.56 & 36.84 \\
\midrule
AdaptSegNet & 45.25 & 35.55 & 35.34 & 21.06 & 36.57 & 34.60 \\ 
FDA & 39.89 & 31.89 & 32.08 & 17.87 & 22.66 & 29.84 \\ 
FADA & 61.00 & 42.19 & 44.23 & 24.57 & 38.15 & 43.10 \\ 
FADA + pAdain & 62.85 & 44.47 & 45.02 & 23.92 & 37.58 & 42.92 \\ 
\midrule
Our Method $\setminus$DC & 58.83 & 42.26 & 42.35 & 24.59 & 39.20 &  41.65 \\

Our Method $\setminus$ST & 62.74 & 43.99 & 44.44 & 24.36 & 41.44 & 43.87  \\

Our Method & {\ul63.41} & {\ul45.99} & {\ul47.28} & {\ul27.37} & {\ul42.03} & {\ul45.71} \\

\bottomrule

 \addlinespace

\end{tabular}
  \caption{Mean intersection over union (mIoU) for the test set of each sub-domain as well as for the whole DRAM dataset. We compare our method to four semantic segmentation domain adaptation methods, and to our method  without the domain confusion step ($\setminus$DC) and without style transfer ($\setminus$ST). Please see details in Section~\ref{sec:experiments}.} 
\label{table: comparison}
\end{center}
\end{table*}

\begin{table*}
\centering
\resizebox{\textwidth}{!}{
\begin{tabular}{@{}c|cccccccccccc|c@{}}
\toprule
Domain             & Background & Bird  & Boat  & Bottle & Cat   & Chair & Cow   & Dog   & Horse & Person & P. Plant & Sheep & mIoU  \\ \midrule
DRAM               & 85.42      & 38.61 & 47.20  & 46.18  & 43.92 & 28.82 & 39.89 & 44.61 & 44.52 & 59.60   & 37.14        & 32.62 & 45.71 \\ \midrule
Realism            & 90.96      & 44.89 & 62.60  & 61.90   & 83.42 & 25.71 & 69.54 & 71.08 & 66.09 & 72.94  & 50.08        & 61.76 & 63.41 \\
Impressionism      & 87.89      & 29.02 & 55.72 & 61.45  & 32.44 & 22.61 & 61.56 & 50.85 & 56.89 & 61.69  & 21.23        & 10.56 & 45.99 \\
Post-Impressionism & 84.26      & 36.69 & 46.81 & 57.95  & 44.24 & 41.47 & 21.00    & 52.51 & 52.76 & 67.94  & 46.80         & 14.92 & 47.28 \\
Expressionism      & 77.93      & 28.45 & 20.70  & 29.63  & 10.55 & 16.38 & 15.10  & 17.08 & 32.91 & 43.05  & 26.23        & 10.49 & 27.37 \\ \midrule
Unseen             & 85.79      & 59.76 & 38.81 & 33.19  & 26.88 & 25.02 & 27.72 & 31.64 & 32.26 & 58.04  & 28.94        & 56.32 & 42.03 \\ \bottomrule
\end{tabular}}
\caption{A breakdown of the mIoU for each class and each movement using our method for semantic segmentation in art paintings. The results can vary significantly also because of class occurrences in a specific movement.} 
\label{table:class results}
\end{table*}

\subsection{Implementation Details}
Our experiments generally follow the hyper-parameters and augmentations strategy suggested by Wang et al.~\cite{fada}. In the following subsection, we elaborate on specific details and additional changes we made.

\textbf{Datasets.} As mentioned in Section~\ref{sec:dataset}, we use PASCAL VOC 2012  \cite{pascal} as the source dataset and DRAM as our target dataset for domain adaptation of semantic segmentation on artistic paintings. PASCAL VOC 2012 is a semantic segmentation dataset containing real images annotated with 20 object classes and one background class. We use the dataset combined with the SBD dataset \cite{sbd} as suggested by Chen et al.~\cite{deeplabv3p}. As we use only 12 classes of the original dataset, we filter out all images which do not contain at least one class of our 11 object classes, resulting in a total of 8362 source annotated images. To equalize the image resolutions, we resize all images in DRAM to have 500 pixels in their largest dimension while keeping original aspect ratio, similarly to PASCAL VOC 2012.


\textbf{Style Transfer Augmentations.} 
We train our style transfer network using PASCAL VOC 2012 as content images and DRAM train set as style images. We use this network to stylize PASCAL VOC 2012 separately for each of DRAM's four sub-domains training data. For each source image we use a single random style image with content/style weight parameter of 0.5. The network was trained on an Nvidia RTX2080Ti GPU for approximately 10 hours.

\textbf{Domain Confusion Network.} We use the FADA~\cite{fada} domain confusion framework with the enhancement of permuted AdaIN layers as presented by Nuriel et al.~\cite{padain}. Similar to previous domain adaptation methods, we use DeepLabV2 \cite{deepLabv2} with Resnet101 \cite{resnet} backbone as the framework's semantic segmentation network. Since we do not use a validation set, we use the same settings used in \cite{fada, padain} for GTA5 -> Cityscapes to train our networks. The only differences are that we use a batch size of 4 instead of 8 because of gpu memory limitations, and we resize images to $513 \times  513$ as suggested by Chen et al.~\cite{deeplabv3p} when training on PASCAL VOC 2012 dataset. As with our style transfer network, we trained the domain confusion step on an Nvidia RTX2080Ti GPU for approximately 10 hours.

\textbf{Baselines and Benchmarks.} Similarly to previous domain adaptation approaches, we use as baseline a DeepLabV2 network trained on PASCAL VOC 2012 and evaluate the results on our DRAM dataset. We also add the results of DeepLabV3+ and HRNet+OCR trained on PASCAL VOC 2012. We compare our adaptation results to the more classic domain adaptation framework AdaptSegNet \cite{adaptsegnet} and to three more recent domain adaptation frameworks: FDA \cite{fda}, FADA \cite{fada} and FADA+pAdain \cite{padain}. We evaluate their results when trained with DRAM dataset as a unified target domain. 
Experimenting with artistic segmentation using multi-domain adaptation frameworks such as OCDA  \cite{openCompoundDA}, and DHA \cite{ocda2} was more challenging as the implementation of these methods is missing. Instead, we used our method with the C-Driving dataset they use and report the results in the supplemental material.
We use the mean intersection over union evaluation method (mIoU) for all experiments. 

\subsection{Results}

Table \ref{table: comparison} summarizes our results. As can be seen, over all sub-domains, as well as on unseen art styles, our method outperforms the alternative methods for semantic segmentation on artistic paintings. 
Table \ref{table:class results} breaks down the semantic segmentation results in art paintings by class and by artistic movement. As can be seen, the availability of class data in a certain art movement can heavily effect the results on a specific class, which in turn can bias the average. For this reason we took care to equalize  class distributions. 



\begin{figure*}
\begin{center}
\includegraphics[width=1.0\textwidth]{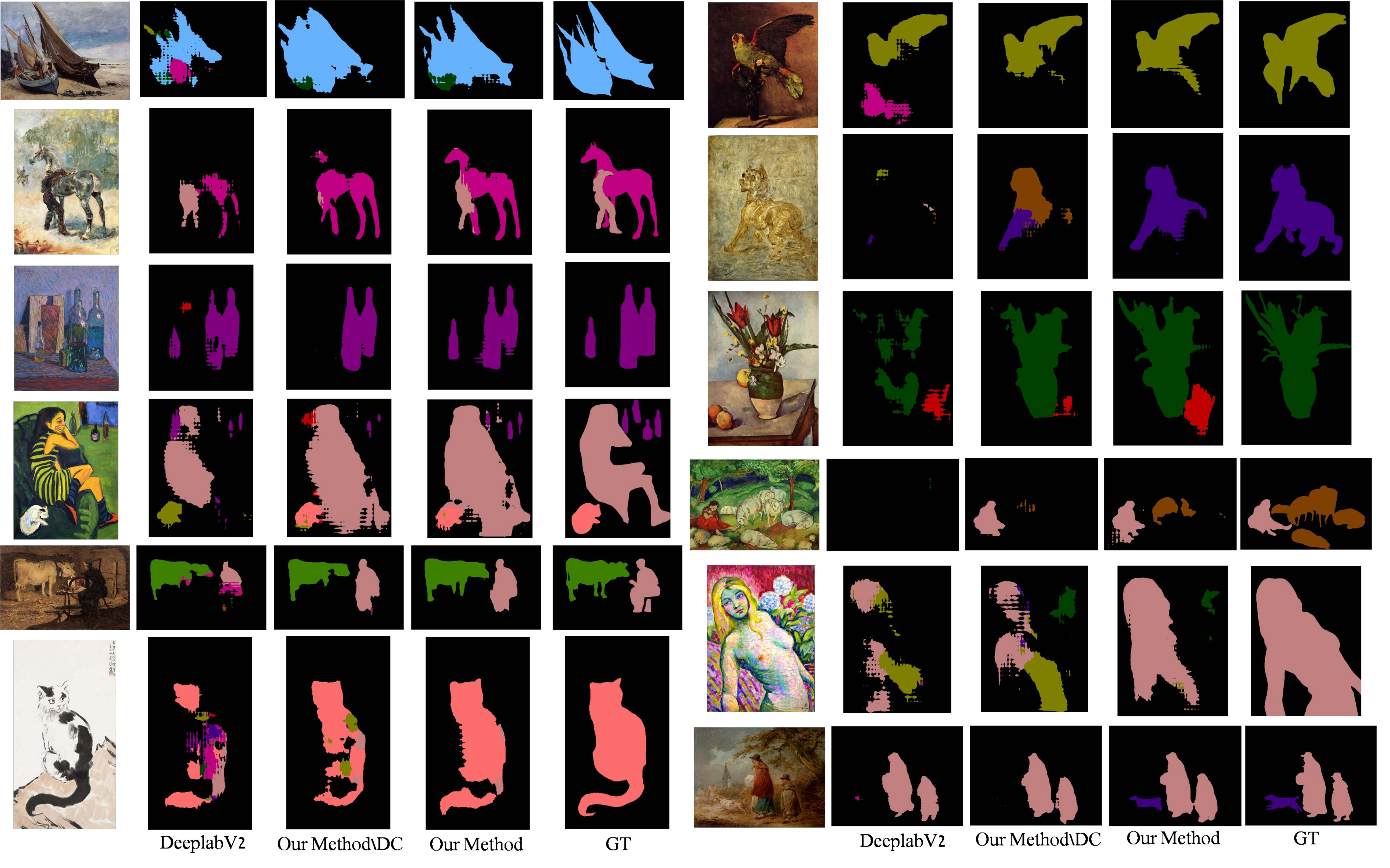}
 \hfill
    \caption{ Qualitative segmentation results on DRAM dataset. The leftmost image is the input image, and the rest (from left to right) present outputs of DeepLabV2 (baseline), Our method$\setminus$DC (only the first step), Our full method, and  ground-truth. Each row corresponds to a different art movement (from top to bottom): Realism, Impressionism, Post-Impressionism, and Expressionism. The last two rows showcase results from the unseen test set. From upper left to bottom right: Divisionism, Fauvism, Ink \& Wash, and Rococo.}
    \label{fig:qualitative}
\end{center}
\end{figure*}

Some qualitative results are shown in Figure~\ref{fig:qualitative}. These demonstrate the improvement gained using our method for all four main art movements and four challenging unseen art movements: Divisionism, Fauvism, Ink \& Wash and Rococo. More segmentation examples can be found in Section~\ref{sec:applications} and in the supplemental materials.

\begin{table*}
\centering
\begin{tabular}{@{}ccccccc|c@{}}
\toprule
Domains                                                                                              & \begin{tabular}[c]{@{}c@{}}Augemntation\\ Style Data\end{tabular} & Realism     & Impressionism & \begin{tabular}[c]{@{}c@{}}Post\\ Impressionism\end{tabular} & Expressionism & Unseen      & DRAM        \\ \midrule
\multicolumn{1}{c|}{\multirow{2}{*}{\begin{tabular}[c]{@{}c@{}}Single Domain\\ (DRAM)\end{tabular}}} & \multicolumn{1}{c|}{No Augmentation}                              & 62.85       & 44.47         & 45.02                                                        & 23.92         & 37.58       & 42.92       \\ \cmidrule(l){2-8} 
\multicolumn{1}{c|}{}                                                                                & \multicolumn{1}{c|}{DRAM}                                         & 63.31       & 43.29         & 43.42                                                        & 21.86         & 32.69       & 40.94       \\ \midrule
\multicolumn{1}{c|}{\multirow{3}{*}{\begin{tabular}[c]{@{}c@{}}Multi \\ Sub-Domains\end{tabular}}}   & \multicolumn{1}{c|}{No Augmentation}                              & 62.74       & 43.99         & 44.44                                                        & 24.36         & 41.44 & 43.87       \\ \cmidrule(l){2-8} 
\multicolumn{1}{c|}{}                                                                                & \multicolumn{1}{c|}{DRAM}                                         & 60.49        & 42.64         & 43.24                                                        & 24.87         & 39.06       & 42.55       \\ \cmidrule(l){2-8} 
\multicolumn{1}{c|}{}                                                                                & \multicolumn{1}{c|}{Target Sub-Domain}                            & {\ul 63.41} & {\ul 45.99}    & {\ul 47.28}                                                  & {\ul 27.37}   & {\ul 42.03}       & {\ul 45.71} \\ \bottomrule
\end{tabular}

\caption{Ablation study for our method using standard segmentation mIoU for evaluations. Results presented on training DRAM as a single domain (Single Domain) or training each sub-domain separately (Multi Sub-Domains). For each of those we evaluate training with no augmentations, source data augmented randomly by DRAM and source data augmented by each sub-domain separately (Target Sub-Domains).  } 
\label{table: ablation}

\end{table*}

\subsection{Ablation Study}

First, as can be seen in Table \ref{table: comparison}, comparing our method with and without the style transfer components (see \textbf{OurMethod$\setminus$DC$\setminus$ST})  shows clearly that augmentation helps close the domain gap. Second, we can see that the domain confusion step improves the results on all artistic movements as well as the unseen ones compared to using only style-transfer augmentation (\textbf{OurMethod$\setminus$DC$\setminus$ST}).

We further study the effect of using our sub-domain training vs.\ training on the full dataset, and using different style-transfer settings for augmentation. We used style transfer with two settings: Using the entire DRAM training set and using each of its sub-domains separately to apply style transfer on the source data. We evaluate the effect of such style transfer when training DRAM as a single domain and when training sub-domain adaptation. The ablation study results can be found in Table \ref{table: ablation}.

As can be observed, using a unique augmented source for training each sub-domain proves to be more beneficial than using a single DRAM-augmented source. Specifically, using separate augmented source datasets with multi-domain adaptation achieves the highest improvement of $2.61\%$ above previous approaches, $8.87\%$ above the highest baseline, and $11.7\%$ above the DeepLabV2 baseline for the entire DRAM test set. In addition, our approach improves each sub-domain result by up to  $2.8\%$ and is especially useful for the more challenging sub-domains. Another important aspect is the superiority of our approach on unseen art styles. Our results clearly show that our inference method achieves better adaptation for unseen art styles -- all multi domain-inference results improve on previous approaches. Our method improves unseen data results by up to 3.88\% above previous approaches, 7.47\% above the highest baseline, and 7.72\% above the DeepLabV2 baseline. This indicates that using style transfer and multi-domain adaptation helps achieve a more general model with better understanding of different art concepts and styles.

Another important aspect presented in Table \ref{table: ablation} is that using the entire DRAM training set for augmentation is not effective. For both domain adaptation and multi-domain adaptation approaches using the entire DRAM training set for stylizing the source dataset causes a significant drop in the results. 
Because of the diversity of the artistic domain represented by DRAM, enabling the network to learn specific features for its different sub-domains results in a significant improvement. 
\section{Applications}
\label{sec:applications}

\begin{figure*}
\begin{center}
\includegraphics[width=1.0\textwidth]{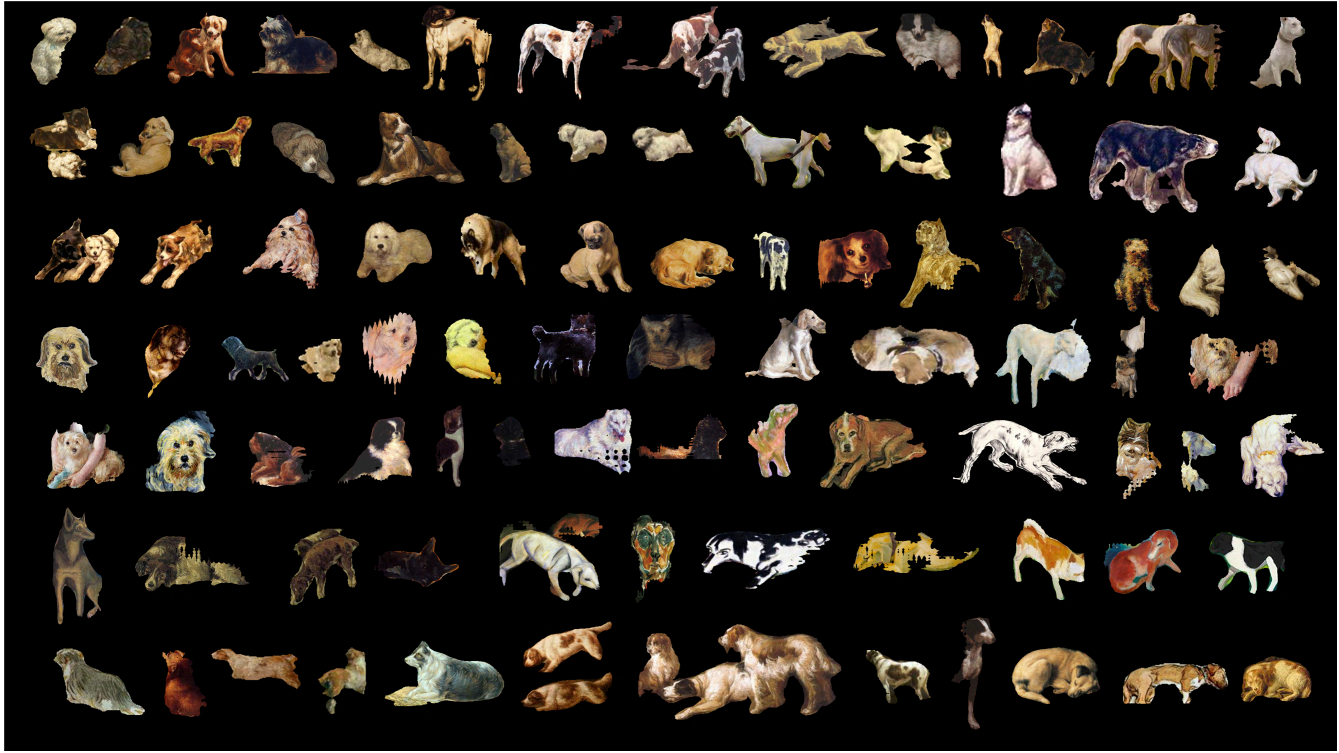}
    \caption{A collection of dog segments from DRAM test set using our method. Such collections allow comparative analysis of artworks.}
    \label{fig: dogs}
\end{center}
\end{figure*}

Semantic segmentation is a fundamental task for understanding and using images in a wide range of applications. We demonstrate two applications of semantic segmentation that can be applied on fine art paintings. The first application focuses on analysis of artwork, and the second on synthesis. 

\subsection{Comparative Collections}

To better understand and analyze artworks in a specific artistic movement or of a specific artist, comparisons are often performed between different paintings. 
Using semantic segmentation, such comparisons can be done not only at the painting level but also on specific objects or items. Using semantic segmentation 
one can gather all occurrences of a certain class from a given set of paintings, extract them from their original images and place them side-by-side for comparison. Figure~\ref{fig: dogs} shows an example of gathering dogs from a set of paintings in the DRAM dataset. To create such a collection, we simply apply our method per painting and create the semantic maps. We then apply connected components labeling \cite{ccl} over all semantic maps. Lastly, we search for images that contain the specific class (Dog) and cut the relevant part out of the original painting.  Such collections can also be used to detect and analyze segmentation errors. On a more abstract level, they depict the perception of a specific concept (Dog) by the network. More collections can be found in the supplemental materials. 

\begin{figure*}
\begin{center}
\includegraphics[width=1.0\textwidth]{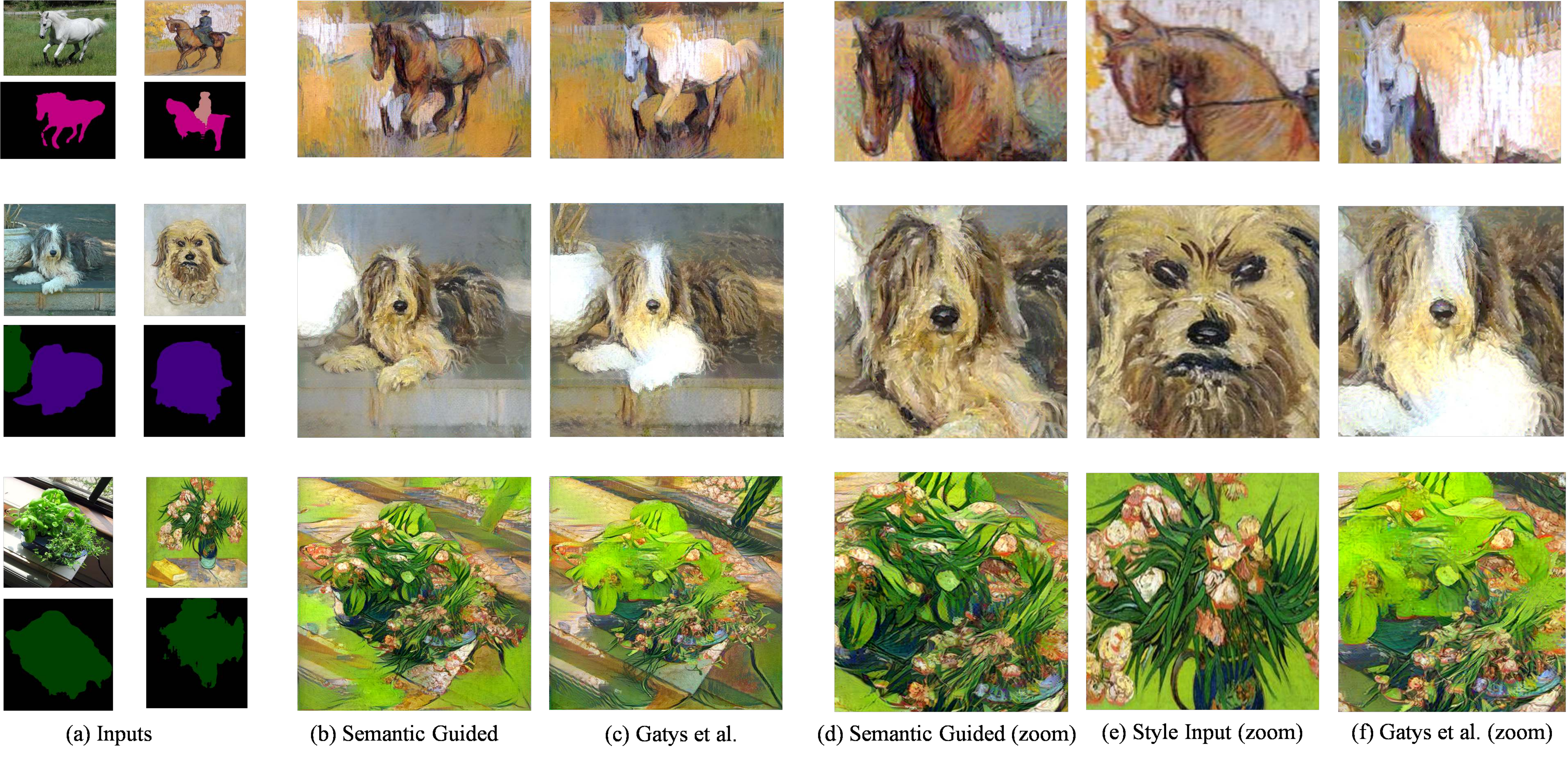}
    \caption{Semantic style transfer examples. The content images were taken from PASCAL VOC 2012 dataset and the style images from the DRAM dataset.     
    (a) shows the automatically segmented network inputs and their corresponding segmentation outputs. (b) shows the result of semantic style transfer and (c) the results of Gatys et al. \cite{gatys2015}. A closer look at specific semantic regions is shown next: (d) is a specific region in the stylized output and (e) is the corresponding semantic region in the style image. Comparison of (d) and (f) shows the same region without semantic stylization.}
    \label{fig: styletransfer}
\end{center}
\end{figure*}

\subsection{Semantic Guided Style Transfer}
Style-transfer, where a photograph is turned into a stylized image based on a chosen stylistic image, has become popular since its introduction in the work of Gatys et al.  \cite{gatys2015}. Most style-transfer methods apply stylization on the whole image (e.g.\ Gatys et al. optimize the style loss over the entire image). However, it may be desirable to break the image to regions, recognize objects and apply a different stylization based on the semantics of the image's content.

Champandard \cite{semanticStyleTransfer} presents a method for semantic style transfer, which requires input of two images (style and content) and two semantic maps (one for each image). The semantic style transfer method encodes the given images to Gram matrices and divides each Gram matrix into patches to find a nearest neighbor style image patch for every content image patch as suggested by Li et al.~\cite{patch_based}. In addition Champandard uses semantic map encodings to add a semantic property to the nearest neighbors patch matching. This encourages the network to use patches taken from the corresponding semantic areas in the optimization process, resulting in a more accurate stylization per specific regions or objects (e.g.\ person->person, horse->horse, see Figure \ref{fig: styletransfer}). 

In its current form, the semantic style transfer method requires manual semantic maps, which can be challenging to create. Using our semantic segmentation method we automate the process and create an end-to-end framework for semantic style transfer that requires only the content/style pair of images. We use our method to create the semantic segmentation map of the style image, and the baseline method for the content image. These maps are used to find common classes and automatically match regions for the semantic style transfer method. 

Figure~\ref{fig: styletransfer} presents some example results. 
As can be observed, the specific style for each common class creates more coherent results, even when the semantic segmentation maps are not perfect.

\section{Discussion}
\label{sec:discussion}







\begin{figure}
\begin{center}
\includegraphics[width=1\columnwidth]{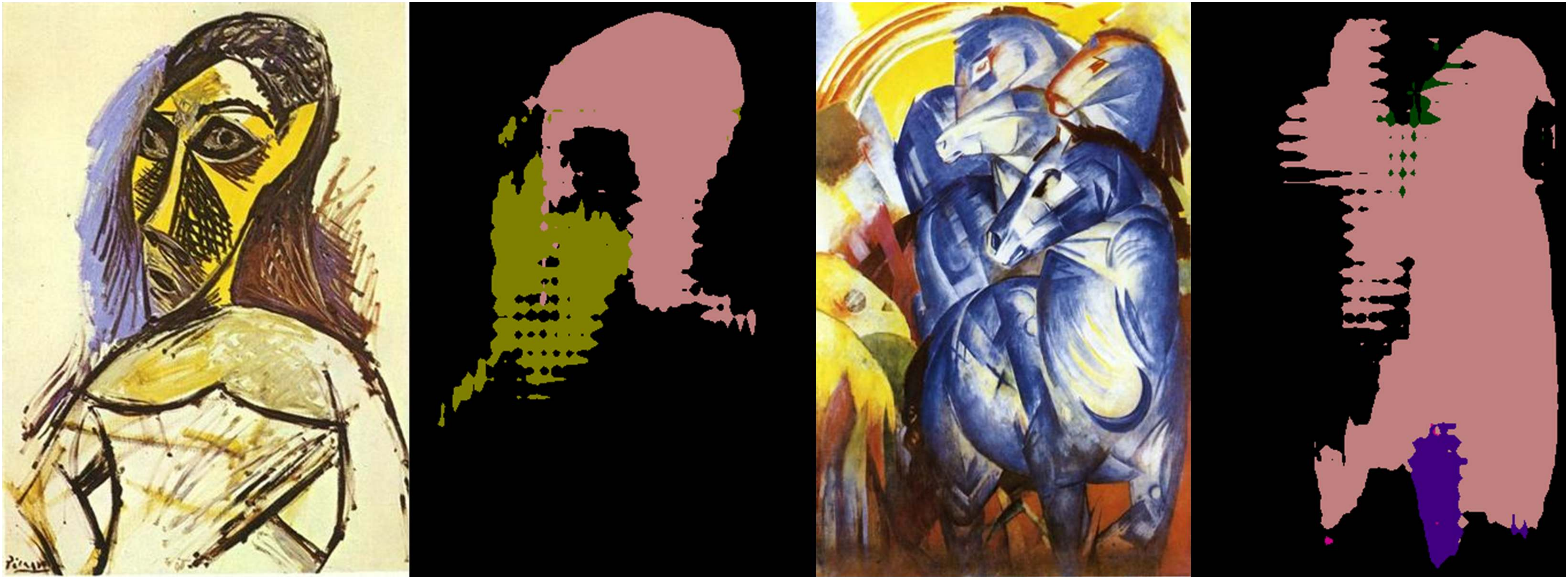}
 \hfill
    \caption{ Examples of results of the Cubism movement (unseen dataset). Although there seems to be a hint of recognition (class recognition on the left and object localization on the right) results are unsatisfactory. We believe that achieving artistic geometric comprehension holds the solution for such geometrically challenging art domains. }
    \label{fig:cubism}
    \vspace{-0.5cm}
\end{center}
\end{figure}

Semantic segmentation is a difficult challenge in general, and more so in the artistic domain. Although our approach provides state-of-the-art results on artistic paintings, there is still a large gap to the ground truth segmentation, and room for improvement in future works. Note that a gap also exists in the results of segmentation of real photographs, although it is smaller. 
In the artistic domain the challenge is greater because of stylization. Our method addresses this by using the Gram-matrices based style feature space. Still, Gram-matrices have a bias towards color and texture. For example, in Figure~\ref{fig: styletransfer}, columns (c) and (f), optimizing the Gram representation of the image without semantic guidance results in output regions which preserve color but fewer brush strokes. 

Painting also involves geometric stylizations. 
In many art movements such as Cubism, Expressionism, and Surrealism, the content of the image may be presented in a geometrically distorted fashion. Human perception can easily understand such paintings and their geometric structure, but this remains a challenge for computer algorithms. Previous methods such as Yaniv et al.~\cite{faceOfArt} achieved better performance by applying geometric augmentation (applying Affine transformations on the images).
We experimented with a variety of geometric augmentation techniques and found that they can assist the results in more stylistic movements such as Post-Impressionism and Expressionism (providing an mIoU of 48.0, 27.92 instead of 47.28, 27.37, respectively). However, since we did not observe an absolute improvement in all artistic movements we decided not to apply geometric stylization by default in our method, and to leave this aspect for future work.

\subsection{Limitations}

Some of our segmentation outputs suffer from artifacts that appear as strokes of circles (see Figure~\ref{fig:cubism}). The reason for this effect is the up-sampling manner of the low dimensional prediction output of DeepLabV2. This happens as small prediction mistakes are exaggerated in the up-sampling process. Another effect caused by up-sampling is missing fine details around edges of objects. Small details disappear when down-sampled in the network, and are thus not considered for prediction. 
To solve these effects one can use larger images, but these may not always be available. DeepLabV3+~\cite{deeplabv3p} offers a decoder module which decreases such effects, and using it along with domain adaptation may provide better results in the future. Additionally, as HRNet+OCR \cite{hrnet, ocr} generalizes better than both DeepLab models over DRAMs unrealistic movements when trained on PASCAL VOC 2012, using it with domain adaptation may also provide an improvement in future results.

Our method holds a few limitations which are derived from the above discussions. 
Since we do not consider geometric aspects to train our method, more abstract art movements have a smaller success rate than realistic ones. For more abstract movements such as Cubism, this can result in unsatisfactory results (Figure \ref{fig:cubism}).
Our method is based on prior knowledge of the movements of the training images to split them to sub-domains. In reality, art data may not hold such information. It may be possible in the future to use style feature spaces to automatically divide an input dataset to sub-domains.

Our dataset holds some limitation related to the complexity of gathering and annotating an art paintings segmentation dataset. As many art movements originated around the 19th century, they do not include modern classes like vehicles and electronic devices which are more common in photographs. This leads our dataset to settle for a relatively small number of classes. Another limitation resulting from annotation complexity is that our dataset lacks a pre-defined validation set. In the future we hope to expand DRAM's class coverage and gather more data as validation.

\subsection{Conclusion}
We presented a first semantic segmentation solution for artistic paintings.
Our unsupervised approach for handling artistic domains achieves state of the art results in comparison to baseline methods for segmentation as well as alternative unsupervised domain adaptation methods. 
We presented the DRAM dataset that includes diverse examples of figurative art paintings and presents a new challenge for domain adaptation because of large domain gaps and large target domain diversity.
We also showed that current domain adaptation approaches that focus on synthetic to realistic driving benchmarks, do not produce the same quality results when trained on realistic to synthetic benchmarks such as ours. 

We defined a composite domain adaptation method that combines sub-domain solutions. We believe this flexible approach can be applied to different kinds of data domains by using better suited augmentation networks and by using different domain adaptation components. 




While art creation applications have developed vastly in recent years, the field of art perception has been less explored. Our work takes a small step towards assisting the art perception of computers. 
We believe that more abstract geometry comprehension is a challenging aspect which may be the key for future advancement. 

\section*{Acknowledgements}
\label{sec:acknowledgements}
This research was partly supported by the Israel Science Foundation (grant No. 1390/19) and The Ministry of Innovation, Science and Technology (grant No. 16470-3).

\printbibliography
\appendix


\begin{small}
\end{small}

\section{Supplemental Materials}

\subsection{DRAM Dataset}

\begin{figure*}[t]
\begin{center}
\includegraphics[width=1.0\textwidth]{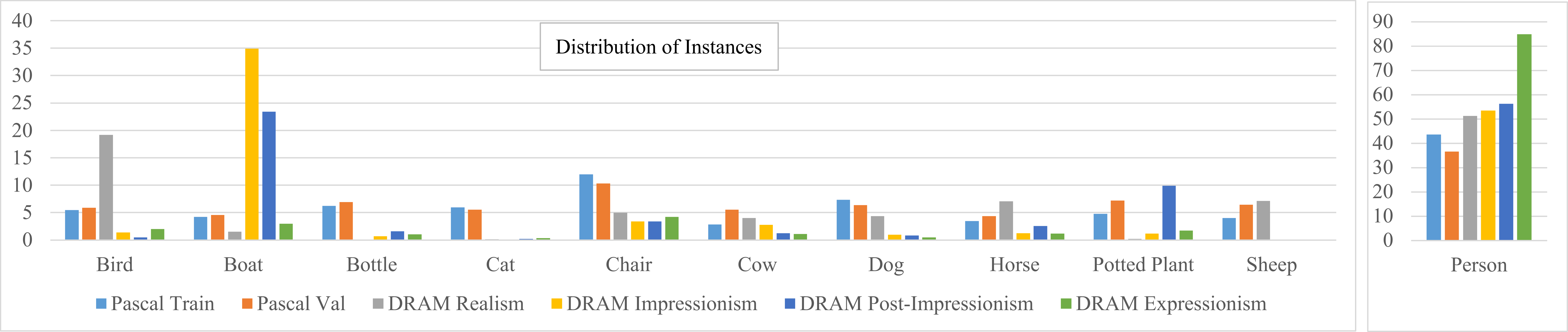}
    \caption{The distribution of number of \emph{instances} for each class in PASCAL VOC 2012 training and validation sets compared to all movements in the DRAM \emph{training set}. As can be seen, we invested effort to equalize these distributions to reduce bias towards any specific class.
}
    \label{fig:set_stats2}
    \vspace{-0.5cm}
\end{center}
\end{figure*}

\subsection{Unseen Art Movements}
An important part of the DRAM dataset is the unseen test set. This set includes 135 images taken from 8 different movements which do not appear in the DRAM training set. Each one of these datasets holds unique characteristics which present new artistic challenges to our method and possibly for future research. In this section we elaborate briefly on our chosen unseen art movements. Examples of paintings from each movement can be found in Figures~\ref{fig:qualitative_art_noveau}
to~\ref{fig:qualitative_rococo}.

\textbf{Art Nouveau}: 
Known as a relatively modern art style which was most popular between 1890 to 1910. It was widely used for interior design and is inspired by natural forms such as plants, animals, and the female figure. One of the main charactaristics of Art-Nouveau is using asymmetric whiplash lines to enhance the sense of dynamism in paintings.

\textbf{Baroque}:
The Baroque art movement flourished in Europe in the 17th century and was strongly encouraged by the Catholic Church to counter the simplicity and austerity of Protestant arts.  Baroque paintings depict dramatic religious scenes.

\textbf{Cubism}: 
An early 20th century art movement. In Cubism, objects are broken up and reassembled in an abstract form which enables the painting to depict the object from multiple viewpoints and represent the subject in greater context. It is considered one of the most influential art movements of the 20th century.

\textbf{Divisionism}: 
Developed around 1884 by applying scientific theories to paintings. Divisionism is defined by seperation of colors to individual dots or patches, believed by divsionist painters to achieve the maximum luminosity scientifically possible (the light emitted from the painting).

\textbf{Fauvism}:
Fauvism is a style known from a French modern artists group named les Fauves ("the wild beasts") and was popular for a short period between 1905 to 1908. Known for preferring unique strong colors over realistic values.

\textbf{Ink \& Wash}: 
A style of East-Asian brush painting which uses the same black ink used in East-Asian calligraphy. It flourished in the Song Dynasty in China (960-1279) and later in Japan. Typically monochrome, with a great emphasis on virtuoso brushwork conveying the perceived spirit of a subject.

\textbf{Japonism}:
Japonism is an art phenomenon from the 19th century following the reopening of trade with Japan in 1858. It was influenced by the rise in popularity of Japanese culture among western artists, who were inspired to create art paintings depicting Japanese styles and culture. 

\textbf{Rococo}: 
Also known as Late Baroque, Rococo emerged in France in the 1730s, and is characterized by a theatrical style combining asymmetry, scrolling curves, and the use of pastels. Rococo images try to depict an illusion of surprise, motion, and drama.  It is often described as the final expression of the Baroque movement.

\subsection{Matching Class Distributions}

To prevent bias in the results, we took steps to equalize the distribution of classes we used from PASCAL VOC 2012 to DRAM dataset. We have shown the distribution of number of pixels in the test set of DRAM to PASCAL, but since the training set is not segmented we manually counted the instances in the images and used this as a measure to equalize the distributions. Figure \ref{fig:set_stats2} presents this distribution. 

\subsection{Implementation Details}
\label{sec:implementation}

\subsection{Style Transfer with AdaIN}
As mentioned in Section~4 in the paper, we use AdaIN style transfer suggested by Huang et al.~\cite{adain-style-transfer} to create pseudo-paintings to train our network in the first step of our method. The AdaIN style transfer approach uses an encoder-decoder architecture connected by an AdaIN layer which aligns the channel-wise mean and variance of a content image to match those of given style image:

\begin{equation}\label{adain_eq}
AdaIN(x_c, x_s) =\sigma(x_s) {x_c - \mu(x_c) \over \sigma(x_c)} + \mu(x_s)
\end{equation}

The encoder used is a pre-trained VGG19 classification network (up to layer relu41), while the decoder replicates the encoder in a reverse fashion and is initialized randomly. The decoder is optimized during the learning process, while the encoder is fixed and is only used for creating the content and style encodings which are fed to the AdaIN layer. The framework uses two loss functions to optimize the decoder, a content loss and a style loss which is weighted by a parameter $\lambda$: 
\begin{equation}\label{content_style_comb}
\ell = \ell_c + \lambda \ell_s
\end{equation}
The content loss is designed to encourage the decoder to produce images with the same statistics as the style image while keeping the details of the content image. It is defined as the euclidean distance between the AdaIN output of a given content/style pair, to the encoding of the output image produced by the decoder:
\begin{equation}\label{adain_content_loss}
\ell_c  = ||f(g(t)) - t||_2 
\end{equation}
where $f(\cdot)$ is the network encoder, $g(\cdot)$ is the network decoder and $t$ is the AdaIN output.

The style loss is designed to encourage similarity between the output and the style image statistics. It is defined as the sum of euclidean distances of the mean and variance of four encoder feature layers (relu11, relu21, relu31, relu41) between the style image and the decoder output:
\begin{equation}\label{adain_stylet_loss}
\begin{split}
\ell_s  = \sum_{i=1}^L {||\mu(\phi_{i}(g(t))) - \mu(\phi_{i}(x_s))||_2} \: + \\
\sum_{i=1}^L {||\sigma(\phi_{i}(g(t))) - \sigma(\phi_{i}(x_s))||_2}
\end{split}
\end{equation}
where $\phi_i$ corresponds to the i'th encoder layer mentioned above.

\subsection{Semantic Segmentation Supervised Training}
Our method uses DeepLabV2 with a Resnet101 backbone as its semantic segmentation network. The network is trained in a supervised fashion with pseudo-paintings (first training step) and with source data images (domain confusion step). The loss function used for optimizing the segmentation network when trained with supervision is the standard cross-entropy loss.

\subsection{Domain Confusion}
Domain confusion adversarial learning generally uses two loss functions. The first is a discrimination loss ($L_D$) which optimizes the discriminator on the loss values generated from domain predictions of both source and target domain inputs. The discriminator's goal is to successfully discriminate between two domains, and $L_D$ is defined as the binary cross entropy function over examples from both source and target domains with labels $(1,0), (0, 1)$, respectively. The second loss is an adversarial loss $L_{adv}$ which optimizes the network's encoder. As the encoder's goal is to confuse the discriminator (aside from producing informative encoding for segmentation); this loss is minimized when the discriminator confuses target samples as coming from the source domain. $L_{adv}$ is also defined as the binary cross entropy function and is evaluated on target examples with the label used for source data for $L_D$.

As described in Section~4 in the paper, our approach for domain confusion is similar to the one presented in FADA \cite{fada} -- Fine Grained Adversarial Learning. The FADA approach utilized class information extracted by the network to enhance the discriminator and create a class-aligned feature domain for the encoding of the segmentation network. To do so, each traditional binary output space of the discriminator is split into k channels (where k is the number of evaluation classes) and the prediction is evaluated for each pixel, resulting in an output of size $H\times W \times 2k$  where $H\times W$ is the original shape of the input image. Instead of the traditional $(0,1), (1, 0)$ discriminator labels, we use the softmax predictions of the segmentation network after truncating the prediction probabilities to $0.9$.  Thus, instead of optimizing on a binary decision between source and target feature space, the discriminator optimizes discriminating every pixel of the image for every class option.
To define $L_D$ and $L_{adv}$ according to the definition above the traditional binary cross entropy function is replaced by averaging the cross entropy over all pixels.  The cross entropy input is the class vectors of size $2k$.

Using these definitions of $L_D$ and $L_{adv}$ the optimization process of the fine-grained adversarial approach can be executed in a manner similar to the more traditional domain confusion described above.

\subsection{Additional Experiments}

\begin{table}

\begin{tabular}{l|ccccc|c}
\toprule
Method & GTA5 -> CityScapes &  CityScapes -> GTA5  \\
\midrule
DeepLabV2 & $37.00$ & $43.71$ \\ 
\midrule
AdaptSegNet & $42.40 (+5.40)$ & $46.52 (+2.81)$  \\ 
FDA & $50.45 (+13.45)$ & $47.32 (+3.61)$  \\ 
FADA & $49.20 (+12.20)$ & $48.32 (+4.61)$   \\
FADA + pAdaIN & $51.50 (+14.50)$ & $48.23(+4.52)$   \\ 
\bottomrule

 \addlinespace

\end{tabular}
  \caption{Mean IOU results of various domain adaptation methods using Cityscapes and GTA5 datasets in both directions. DeepLabV2 \cite{deepLabv2} is used as the baseline - one can observe that the gain in using adaptation is much higher for the direction GTA5 to CityScapes. } 
\label{tab:city2gta}
\end{table}
\subsection{Adaptation Direction}

\begin{table*}
\centering
\begin{tabular}{@{}c|ccccc|c@{}}
\toprule
Method                         & Realism     & Impressionism & Post-Impressionism & Expressionism & Unseen     & DRAM  \\ \midrule
FADA (Step1)        & 52.45       & 36.26         & 30.57              & 15.22         & 34.31      & 34.01 \\
FADA+pAdaIN (Step1) & 54.87       & 37.59         & 32.95              & 17.58         & 35.64      & 35.98 \\ \midrule
Our Method \textbackslash{}DC  & {\ul 58.83} & {\ul 42.26}   & {\ul 42.35}        & {\ul 24.59}   & {\ul 39.20} & {\ul 41.65} \\ \bottomrule
\end{tabular}
\caption{Mean Intersection over Union (mIoU) results for the supervised learning step of FADA, FADA+pAdaIN compared to our method. Note that Our Method\textbackslash ST uses the same first step as FADA+pAdaIN as it uses a single source domain.}
\label{tab:step1_results}
\end{table*}

Most domain adaptation approaches present results on adapting synthetic source data to realistic target data (e.g. GTA5 \cite{gta5} to CityScapes \cite{cityscapes}). We seek an opposing direction -- training with a photographic source dataset and adapting to a synthetic target dataset of fine art paintings. 
We compared different domain adaptation frameworks by evaluating them in the opposite direction that they were designed for: using CityScapes as the source dataset and GTA5 as the target dataset.
We chose to use these datasets (although in reverse) because they are typically used for benchmarks in most domain adaptations for semantic segmentation approaches. 
This experiment is important as it emphasizes the difficulty increase when training a realistic to synthetic DA task rather than the usual synthetic to realistic DA task.
We also used this experiment to choose a backbone DA framework for our task. 

As can be observed in Table \ref{tab:city2gta}, all methods struggle to reproduce results with the same quality as the original adaptation direction. 
This experiment clearly shows that domain adaptation is not symmetric. The reduction in the results may be due to using a network that is pretrained on realistic datasets such as  ImageNet \cite{imagenet} for adaptation, or simply a result of overfitting to the specific problem of GTA5 to CityScapes. This demonstrates the need to find a better solution for adapting real to synthetic domains.

\subsection{Step1 Comparison}

Our proposed method is comprised of two main steps, pseudo-paintings supervised-learning and domain confusion. To further emphasis the effect of using pseudo-paintings, we present results for the initial supervised steps of FADA and FADA+pAdaIN in comparison to our first step (see Table~\ref{tab:step1_results}). These results show a clear advantage for using pseudo-paintings to compensate for the large domain gap between PASCAL VOC 2012 to DRAM. Additionally, it shows the importance of using pAdaIN regularization in order to further compensate for the large domain gap as FADA+pAdaIN achieves higher results than FADA on all art movements.

\subsection{Multi-Target Comparisons}
We compare our method to the Multi-Target DA methods OCDA \cite{openCompoundDA} and DHA  \cite{ocda2}. As it was not possible to utilize our data on their Multi-Target solutions (no code or explanation on the semantic segmentation part of their method was available), we chose to use the C-Driving dataset \cite{openCompoundDA} that they use as benchmark.    
To make as equitable comparison as possible we used our method with a VGG16 \cite{vgg16} backbone as used in OCDA and DHA. 
As can be observed in Table \ref{tab:cdriving_results} our method achieves a 5.41\% increase over OCDA on average, but is not superior to DHA method.
However, our network is designed for semantic segmentation of artistic paintings, not C-Driving. We believe that replacing the Gatys style representation~\cite{gatys2015} and the AdaIN style transfer \cite{adain-style-transfer} with a better suited transformation model could provide better results but it is not the focus of our work. 

\begin{table}
\begin{tabular}{@{}c|cccc|c@{}}
\toprule
Method       & Rainy     & Snowy   	  & Cloudy         & Overcast   & Avg.\\ \midrule
OCDA        & 22.00         & 22.90           & 27.00                & 27.90           & 25.00   \\
DHA 		  & {\ul 27.10}       & {\ul 30.40}         	 &{\ul 35.50}           & {\ul 36.10}     & {\ul 32.30}  \\ \midrule
Our Method  & 26.15    & 26.42        & 31.95              & 32.49 			& 30.41 \\ \bottomrule
\end{tabular}
\caption{Mean Intersection over Union (mIoU) results for OCDA and DHA compared to our method on the C-Driving benchmark.}
\label{tab:cdriving_results}
\end{table}

\subsection{Additional Results}
\begin{figure*}
\begin{center}
\includegraphics[width=1.0\textwidth]{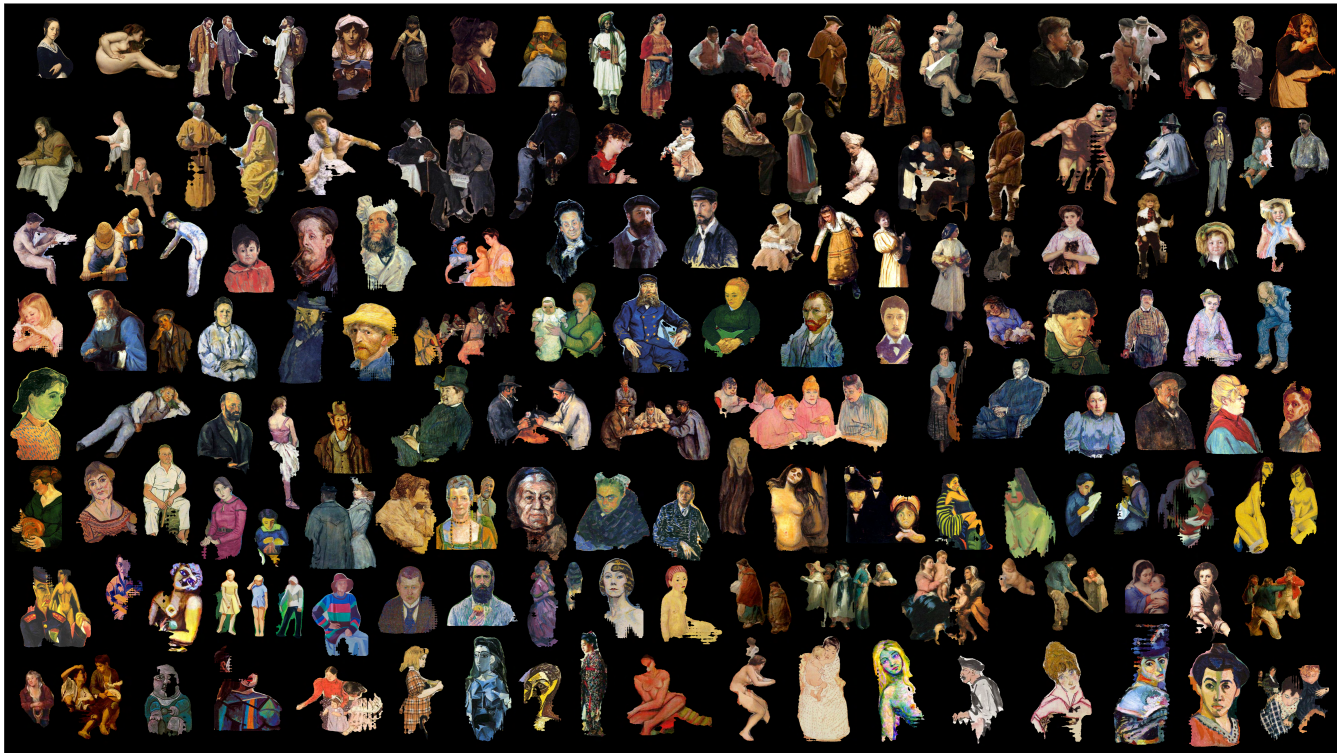}
    \caption{A collection of human figure (person) segments from the DRAM test set extracted using our method.}
    \label{fig: person}
\end{center}
\end{figure*}

\begin{figure*}
\begin{center}
\includegraphics[width=1.0\textwidth]{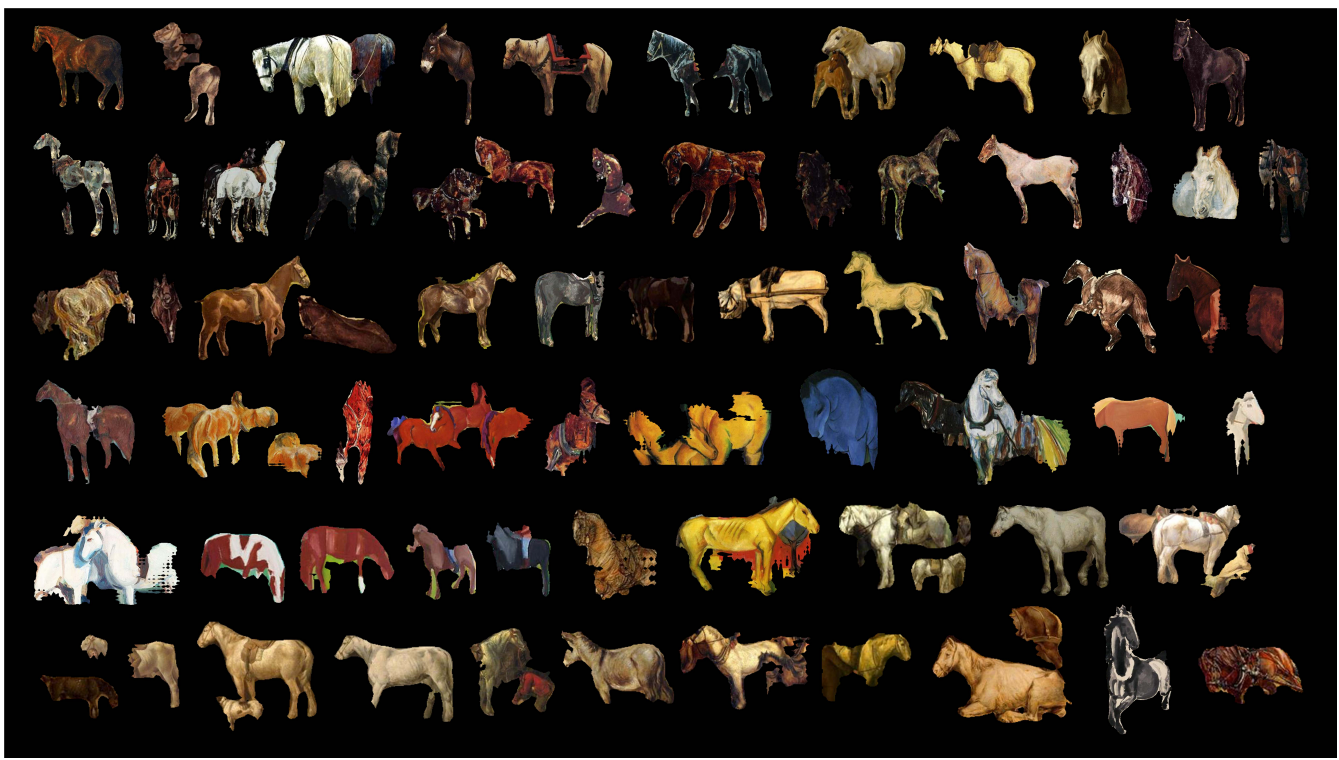}
    \caption{A collection of horse segments from DRAM test set extracted using our method.}
    \label{fig: horses}
\end{center}
\end{figure*}

\begin{figure*}
\begin{center}
\includegraphics[width=1.0\textwidth]{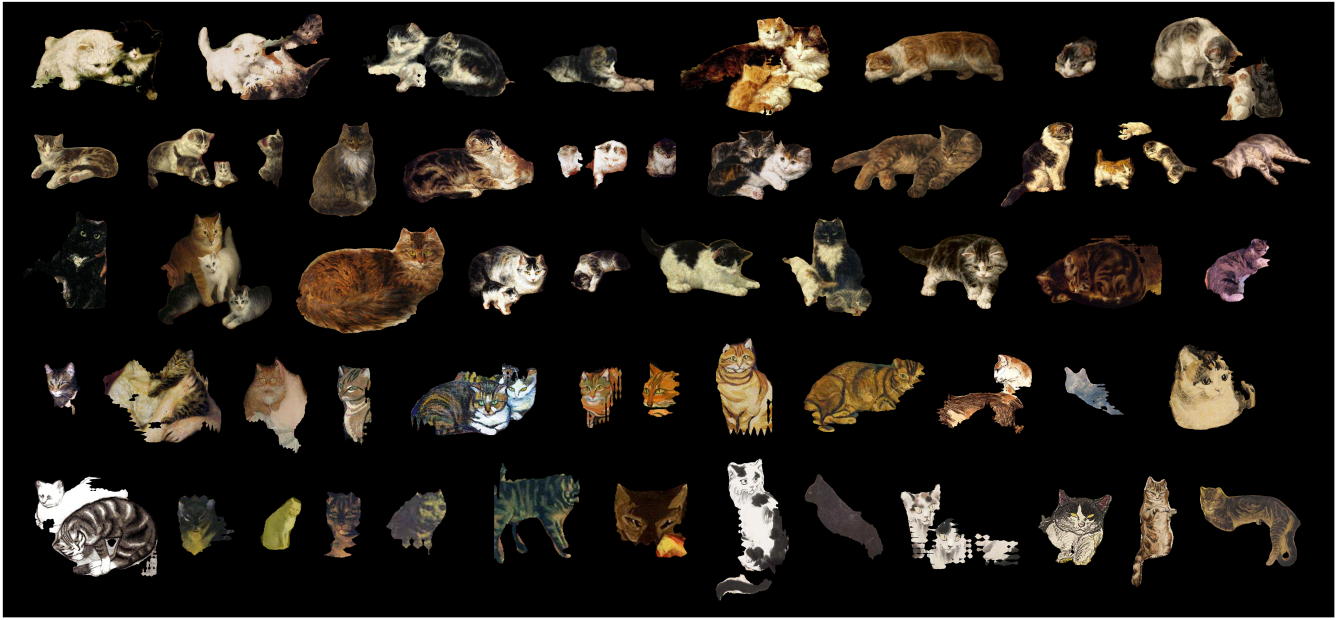}
    \caption{A collection of cat segments from DRAM test set extracted using our method.}
    \label{fig: cats}
\end{center}
\end{figure*}

\begin{figure*}
\begin{center}
\includegraphics[width=1.0\textwidth]{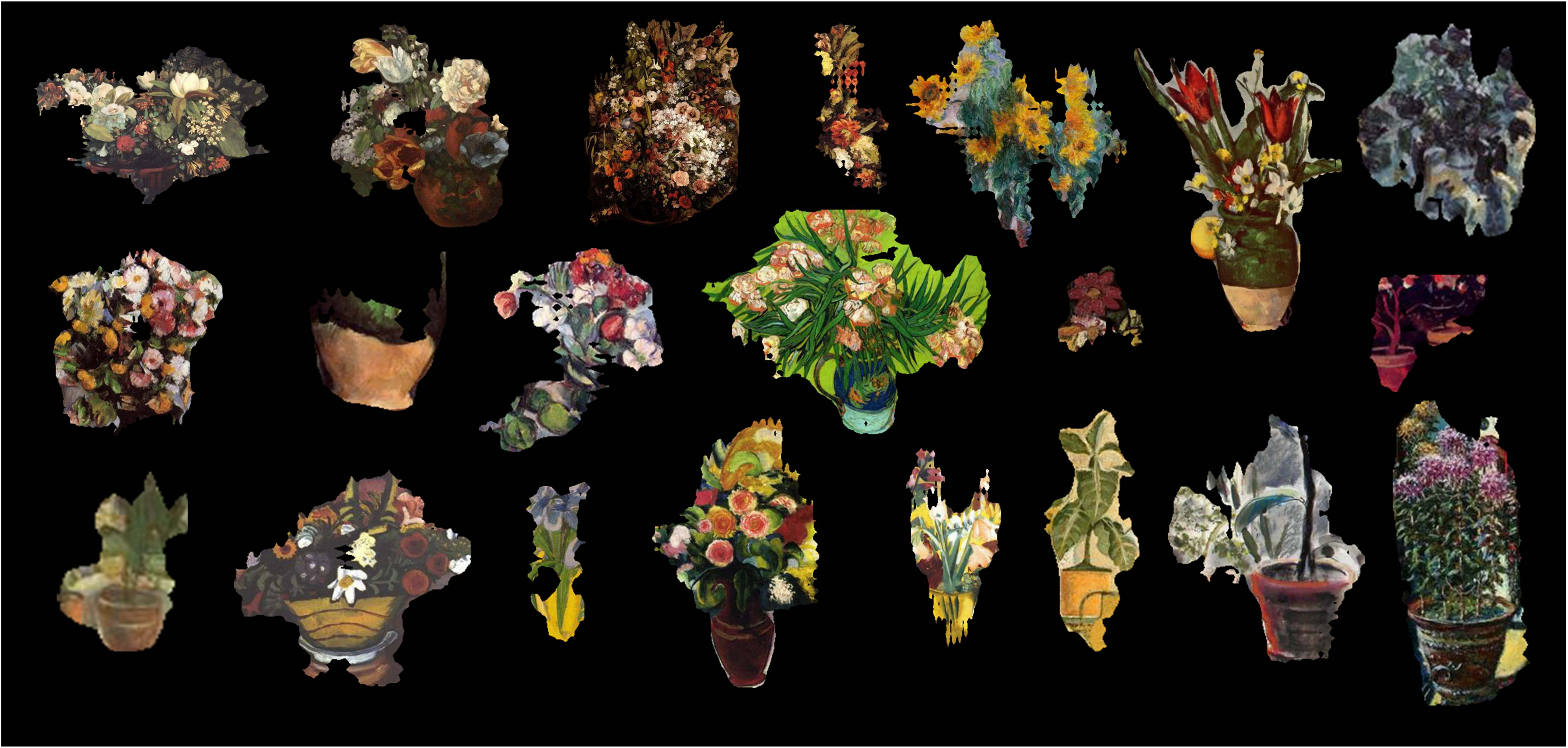}
    \caption{A collection of potted plant segments from DRAM test set extracted using our method.}
    \label{fig: potted plant}
\end{center}
\end{figure*}

Figures~\ref{fig: person} to~\ref{fig: potted plant} show additional examples of comparative collections created for Person, Horse, Cat, Potted Plant classes from the DRAM test set.

Figures~\ref{fig:qualitative_realism} to~\ref{fig:qualitative_expressionism} show additional qualitative examples of our segmentation results on the four sub-domains that are used for training in DRAM. Figures~\ref{fig:qualitative_art_noveau}
to~\ref{fig:qualitative_rococo} present additional qualitative example of our segmentation results on the additional unseen domains in DRAM test set. In all qualitative example figures the leftmost image is the input image, and then (from left to right) the results of DeepLabV2 (baseline), our method$\setminus$DC (only the first step), our full method, and then ground-truth.



 

\section{Appendix - List of Paintings}
\label{sec:appendix}
\subsection{Main Paper}
{\ul Figure~\ref{fig:teaser} from top left:} \\
Printemps. Charles Jacque. Wikimedia. Public Domain. \\
Fishing Boats, Calm Sea. Claude-Monet. Wikiart. Public Domain. \\
Grazing Horses. Franz Marc. Wikiart. Public Domain. \\
The Scream. Edvard Munch. Wikiart. Public Domain.\\
Cows in the Field. Constant Troyon. Wikiart. Public Domain.\\
Dun, a Gordon Setter Belonging to Comte Alphonse de Toulouse Lautrec.  Henri de-Toulouse Lautrec. Wikiart. Public Domain.\\
At the Races. Henri de-Toulouse Lautrec. Wikiart. Public Domain.\\
Two Yellow Knots with Bunch of Flowers. Ernst Ludwig Kirchner. Wikiart. Public Domain.\\
Simpkin at the Tailor’s Bedside. Beatrix Potter. Wikiart. Public domain.\\
The Green Line. Henri Matisse. Wikiart. Public Domain US.\\
Three Ducks. Xu Beihong. Wikiart. Public domain China.\\
Nature Morte Cubist. Louis Marcoussis. Wikiart. Public Domain.\\
Japanese. Vasily Vereshchagin. Wikiart. Public Domain.\\
\\
{\ul Figure~\ref{fig:overview}:}\\
Realism Batch:\\
A Greenland, or Gyr Falcon. Archibald Thornburn. Wikiart. Public Domain.\\
A bouquet of flowers. Ilya Repin. Wikiart. Public Domain.\\
A Fisher Girl. Ilya Repin. Wikiart. Public Domain.\\
The Return from the Mill. Rosa Bonheur. Wikiart. Public Domain.\\
Brizo, a Shepherd's Dog. Rosa Bonheur. Wikiart. Public Domain.\\
Portrait of Lucy Langdon Williams Wilson. Thomas Eakins. Wikiart. Public Domain.\\
Gloucester Harbor. Winslow Homer. Wikiart. Public Domain.\\
Sheep on the Downs. James Ward. Wikiart. Public Domain.\\
Impressionism Batch:\\
Berck: Low Tide. Eugene Boudin. Wikiart. Public Domain.	\\	
Harnessed Horses. Eugene Boudin. Wikiart. Public Domain.\\		
Mother and Child. John Henry Twachtman. Wikiart. Public Domain.\\
The red blouse. Berthe Morisot. Wikiart. Public Domain.	\\	
The Cage. Berthe Morisot. Wikiart. Public Domain.\\	
Julie Manet and her Greyhound Laerte. Berthe Morisot. Wikiart. Public Domain.\\	
On the Beach. Eduard Manet. Wikiart. Public Domain.\\	
Lilac in a glass. Eduard-Manet. Wikiart. Public Domain.\\
Post-Impressionism Batch:\\
Head of Lorette with Curls. Henri Matisse. Wikiart. Public Domain.\\		
Evening in a Russian Village. Konstantine Ivanovich Gorbatov. Wikiart. Public Domain.\\		
Woman with necklace of gems. Pablo Picasso. Wikiart. Public Domain.\\		
The picador. Pablo Picasso. Wikiart. Public Domain.\\		
Arearea I. Paul Gauguin. Wikiart. Public Domain.\\		
Self Portrait with mandolin. Paul Gauguin. Wikiart. Public Domain.\\	
Aspidistra. Samuel Peploe. Wikiart. Public Domain.\\		
Vase of Flowers. Eduard Vuillard. Wikiart. Public Domain.\\
Expressionism Batch:\\
Green Eye Mask. Amadeo De-Souza Cardoso. Wikiart. Public Domain.\\	
Portrait of Francisco Cardoso. Amadeo De-Souza Cardoso. Wikiart. Public Domain.\\
The Greyhounds. Amadeo De-Souza Cardoso. Wikiart. Public Domain.\\
Girl. Sitting Female Nude. Max Pechstein. Wikiart. Public Domain.\\	
The Masked Woman. Max Pechstein. Wikiart. Public Domain.\\	
L'artiste et sa Femme. Gustave De-Smet. Wikiart. Public Domain.\\
La vie du Ferme. Gustave De-Smet. Wikiart. Public Domain.\\
Leopold Zborowski. Amedeo Modigliani. Wikiart. Public Domain.\\
\\
{\ul Figure~\ref{fig:pseudo_paintings}:} \\
Girl with Flowers. Daughter of the Artist. Ilya Repin. Wikiart. Public Domain. \\
Winter (aka Woman with a Muff). Berthe Morisot. Wikiart. Public Domain. \\
Boats by the River Bank. Konstantin Gorbatov. Wikiart. Public Domain. \\
The Pagans. Oskar Kokoschka. Wikiart. Public Domain US.\\
Martin, a Terrier. Rosa Bonheur. Wikiart. Public Domain.\\
The Bridge over the Toques at Deauville. Eugene-Boudin. Wikiart. Public Domain.\\
Royan, Charente Inferieure. Samuel Peploe. Wikiart. Public Domain.\\
The Carnival. Paula Modersohn Becker. Wikiart. Public Domain.\\
\\
{\ul Figure~\ref{fig:inference}:} \\
Cormorant. Xu Beihong. Wikiart. Public Domain China. \\
\\
{\ul Figure~\ref{fig:qualitative}} \\
Fishing Boats on the Deauville Beach. Gustave Courbet. Wikiart. Public Domain.\\
The Green Parrot. Vincent Van-Gogh. Wikiart. Public Domain.\\
Artilleryman Saddling His Horse. Henri De-Toulouse Lautrec. Wikiart. Public Domain.\\
The Dog (Sketch of Touc). Henri De-Toulouse Lautrec. Wikiart. Public Domain.\\
Still Life with Bottles, Roderic O'Conor. Wikiart. Public Domain.\\
Still Life, Tulips and apples. Paul Cezanne. Wikiart. Public Domain.\\
Female Artist. Ernst Ludwig Kirchner. Wikiart. Public Domain.\\
Shepherdess with Sheep. Franz Marc. Wikioo. Public Domain.\\
Spinning. Giovanni Segantini. Wikiart. Public Domain.\\
Nu (Nude). Jean Metzinger. Wikiart. Public Domain US.\\
Cat. Xu Beihong. Wikiart. Public Domain China.\\
Woman, Child and Dog on a Road. George Morland. Wikiart. Public Domain.\\
\\
{\ul Figure~\ref{fig: styletransfer}} \\
Amazone. Henri De-Toulouse Lautrec. Wikiart. Public Domain. \\
Head of the Dog. Claude Monet. Wikiart. Public Domain.\\
Oleanders and Books. Vincent Van-Gogh, Wikiart. Public Domain.\\
\\
{\ul Figure~\ref{fig:cubism}} \\
Female nude (study). Pablo Picasso. Wikiart. Public Domain US. \\
The Tower of Blue Horses. Franz Marc. Wikiart. Public Domain.

\subsection{Supplemental Materials}
\begin{figure*}[t]
\begin{center}
\includegraphics[width=1.0\textwidth]{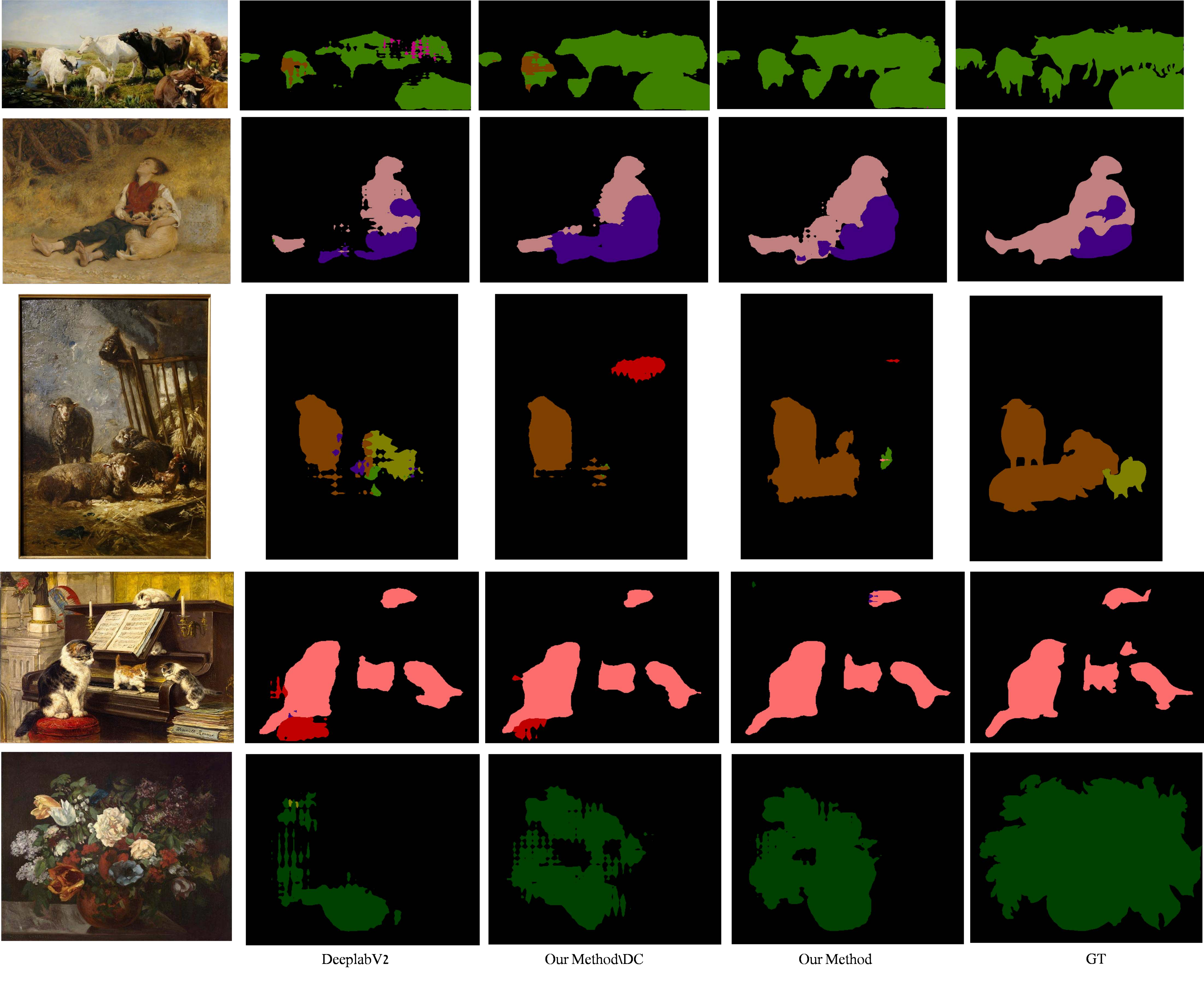}
 \hfill
    \caption{ Qualitative segmentation results from DRAM Realism test set.}
    \label{fig:qualitative_realism}
\end{center}
\end{figure*}

\begin{figure*}[t]
\begin{center}
\includegraphics[width=1.0\textwidth]{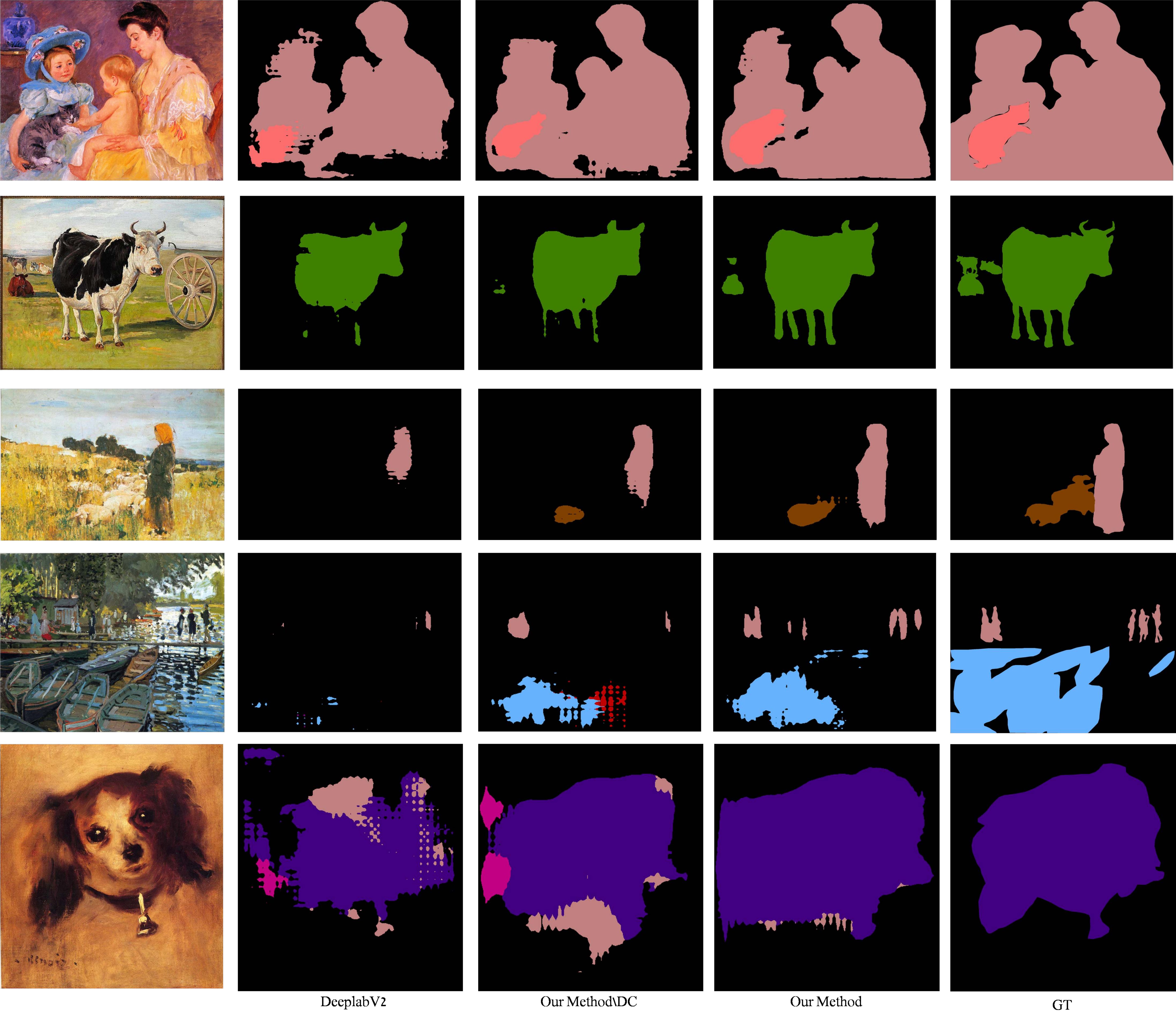}
 \hfill
    \caption{ Qualitative segmentation results from DRAM Impressionism test set.}
    \label{fig:qualitative_impressionism}
\end{center}
\end{figure*}

\begin{figure*}[t]
\begin{center}
\includegraphics[width=1.0\textwidth]{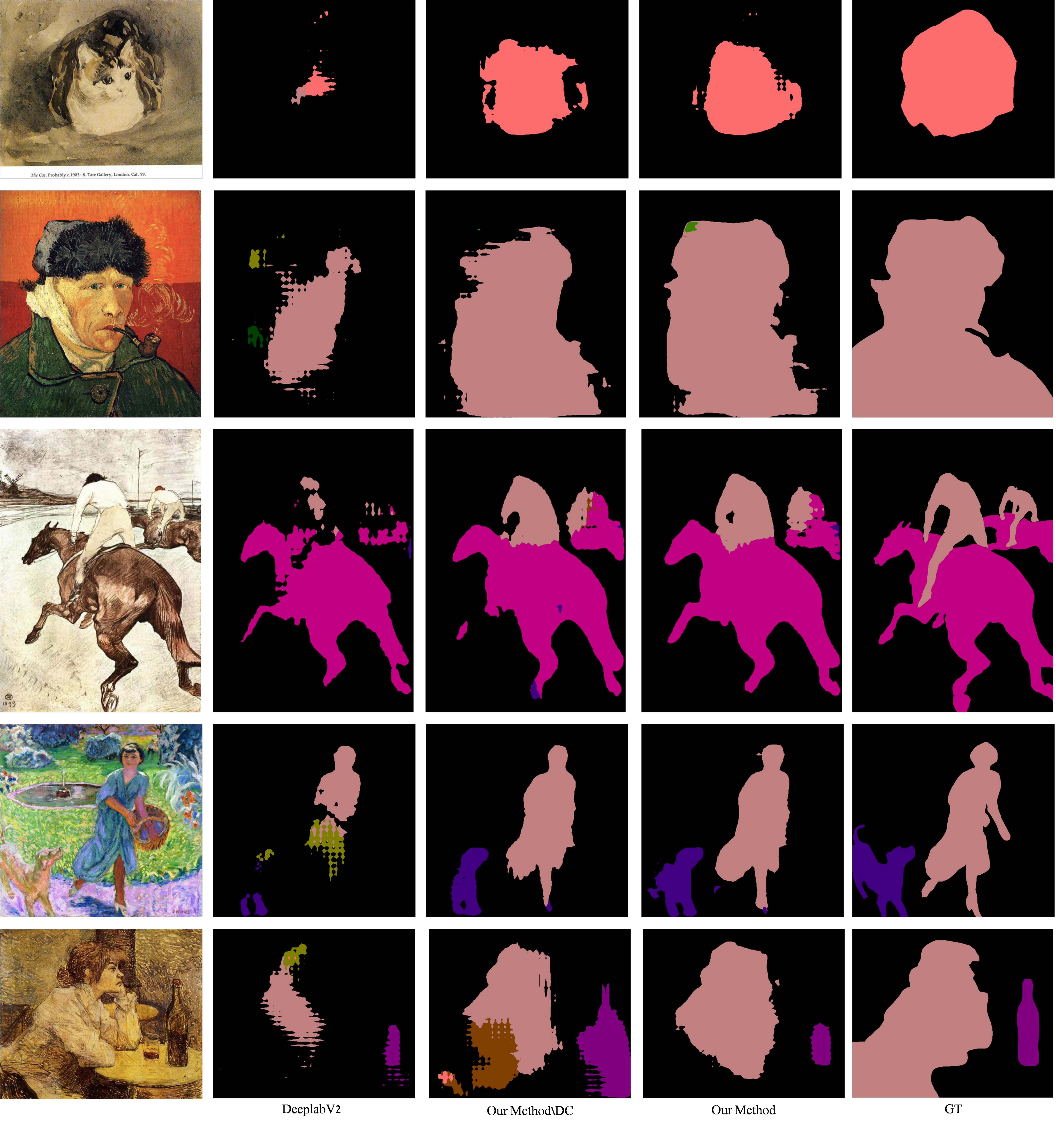}
 \hfill
    \caption{ Qualitative segmentation results from DRAM Post-Impressionism test set.}
    \label{fig:qualitative_post_impressionism}
\end{center}
\end{figure*}

\begin{figure*}[t]
\begin{center}
\includegraphics[width=1.0\textwidth]{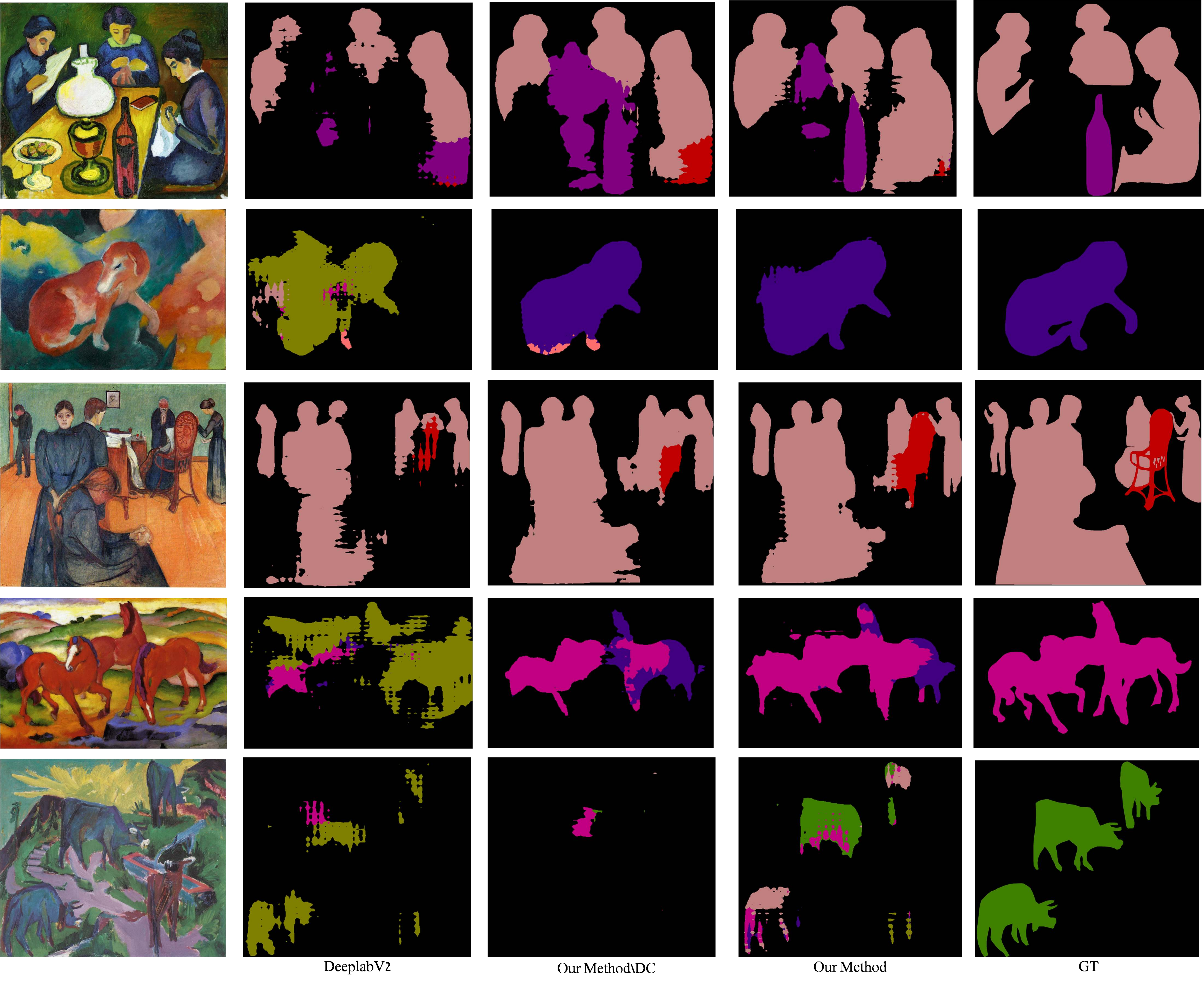}
 \hfill
    \caption{ Qualitative segmentation results from DRAM Expressionism test set.}
    \label{fig:qualitative_expressionism}
\end{center}
\end{figure*}

\begin{figure*}[t]
\begin{center}
\includegraphics[width=1.0\textwidth]{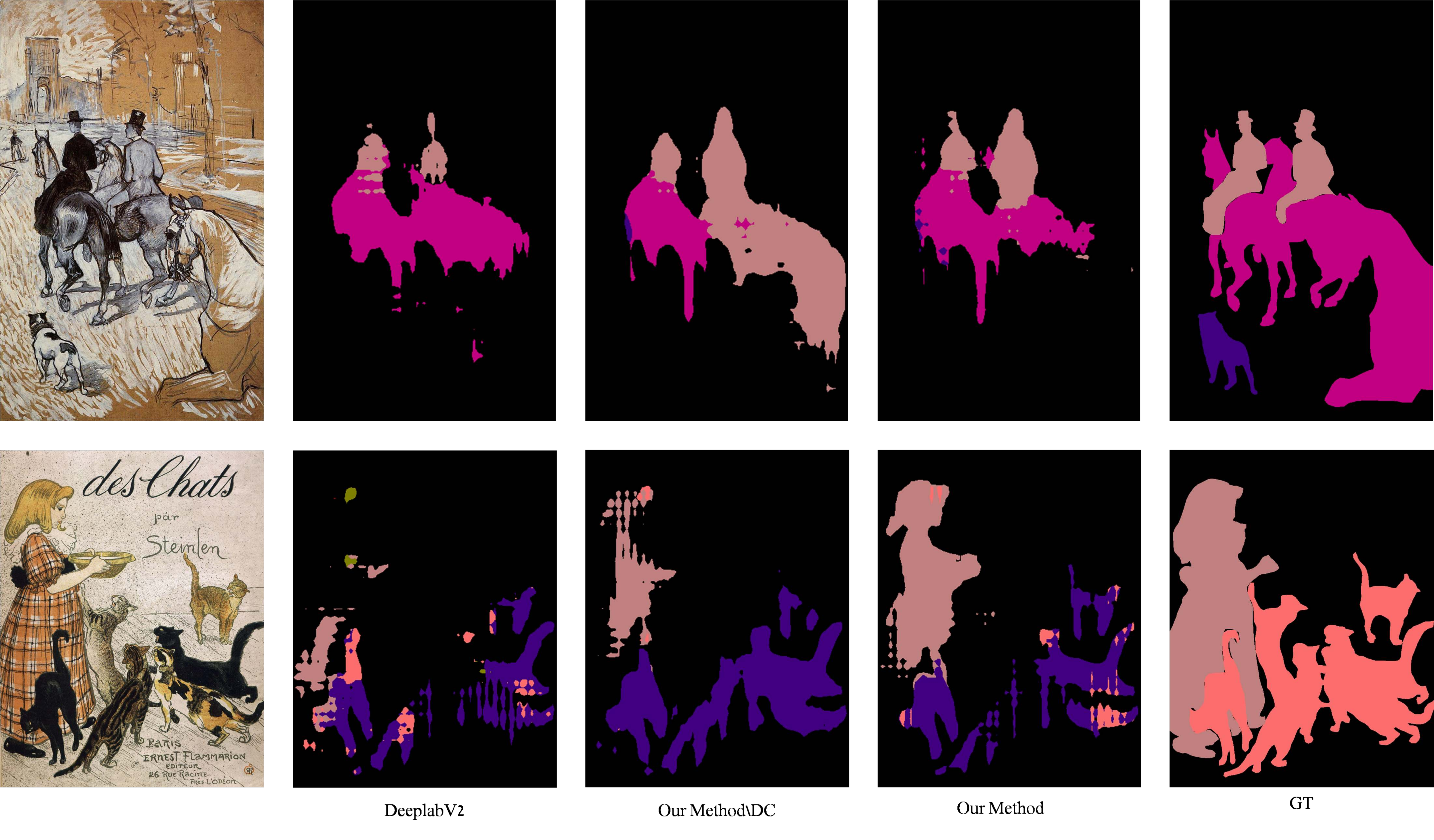}
 \hfill
    \caption{ Qualitative segmentation results from DRAM Art-Nouveau unseen test set.}
    \label{fig:qualitative_art_noveau}
\end{center}
\end{figure*}

\begin{figure*}[t]
\begin{center}
\includegraphics[width=1.0\textwidth]{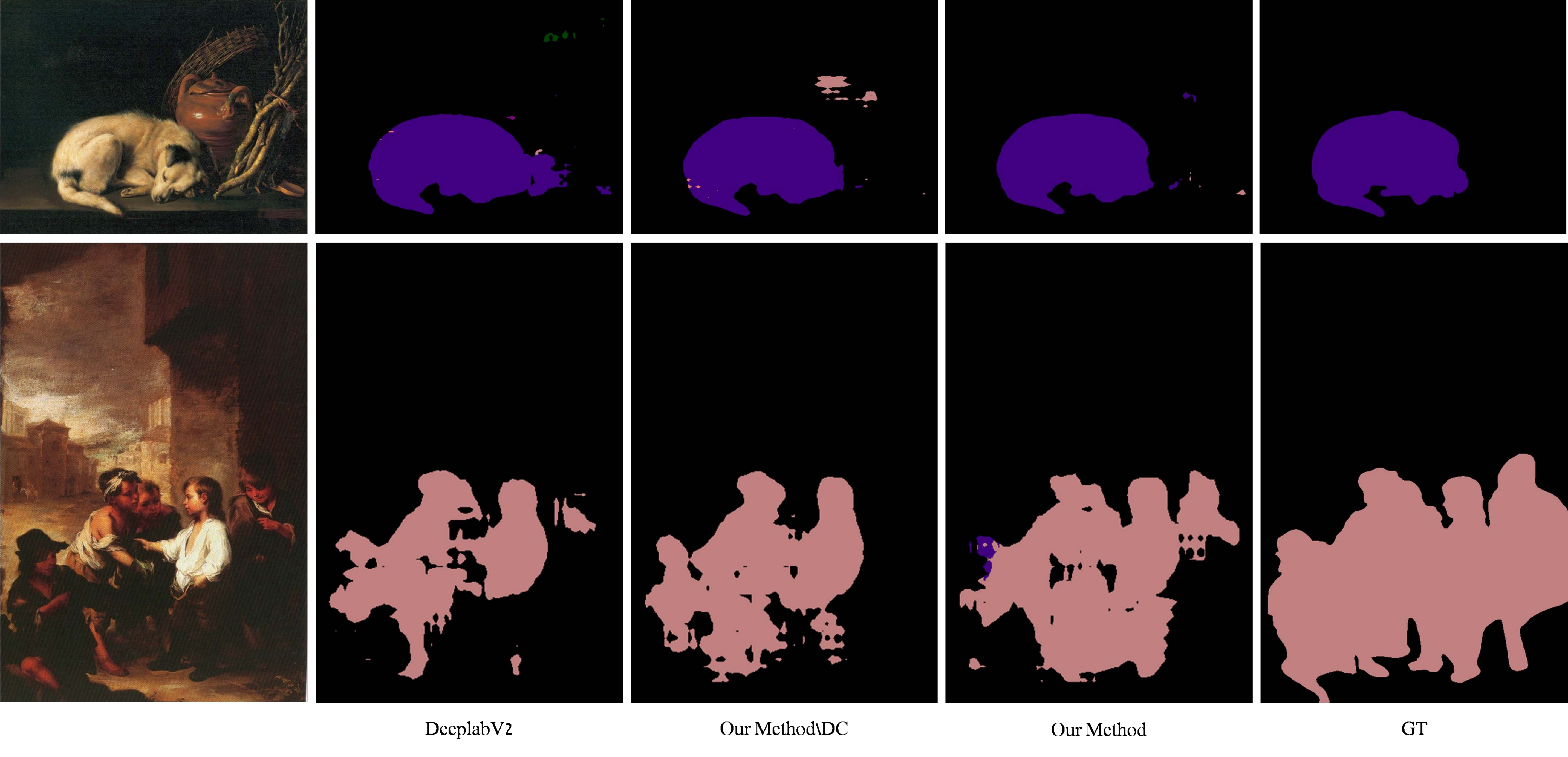}
 \hfill
    \caption{ Qualitative segmentation results from DRAM Baroque unseen test set.}
    \label{fig:qualitative_baroque}
\end{center}
\end{figure*}

\begin{figure*}[t]
\begin{center}
\includegraphics[width=1.0\textwidth]{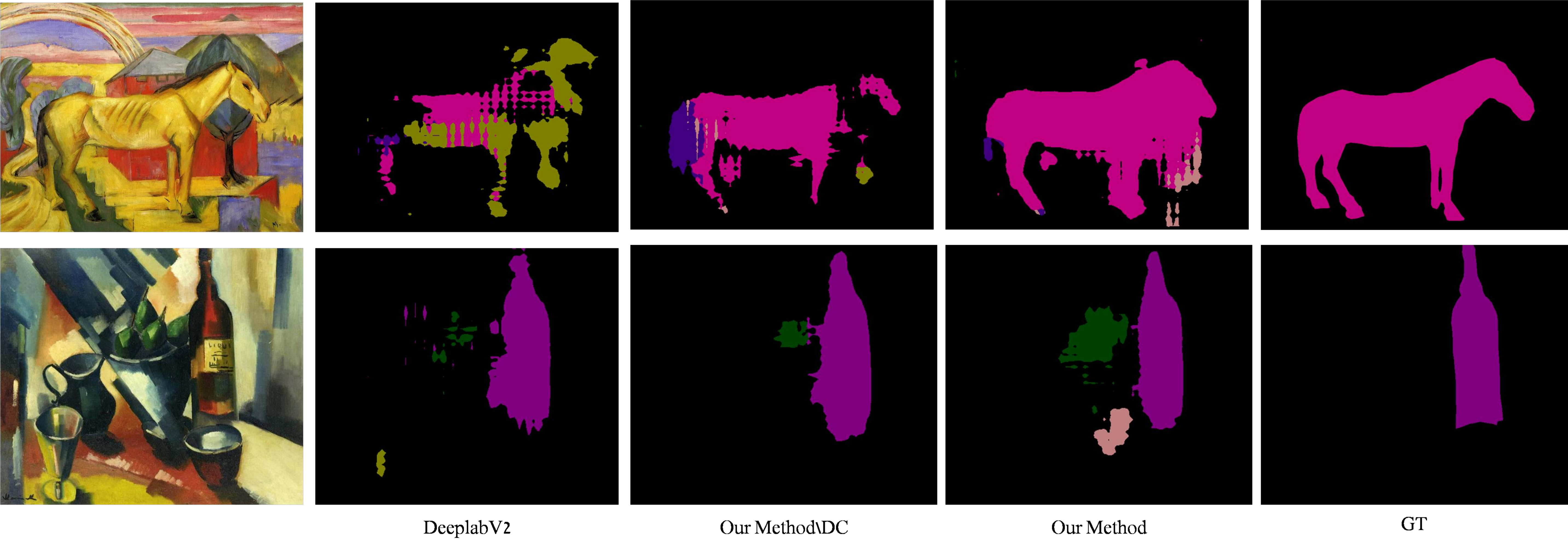}
 \hfill
    \caption{ Qualitative segmentation results from DRAM Cubism unseen test set.}
    \label{fig:qualitative_cubism}
\end{center}
\end{figure*}

\begin{figure*}[t]
\begin{center}
\includegraphics[width=1.0\textwidth]{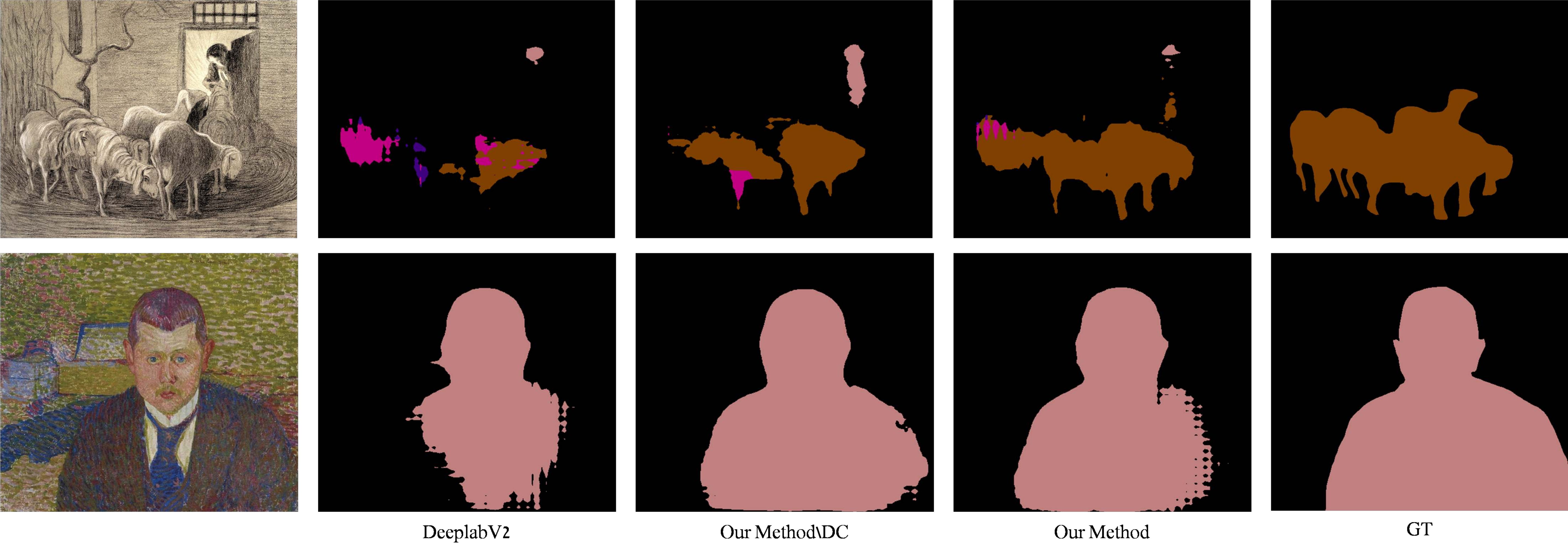}
 \hfill
    \caption{ Qualitative segmentation results from DRAM Divisionism unseen test set.}
    \label{fig:qualitative_divisiosnim}
\end{center}
\end{figure*}

\begin{figure*}[t]
\begin{center}
\includegraphics[width=1.0\textwidth]{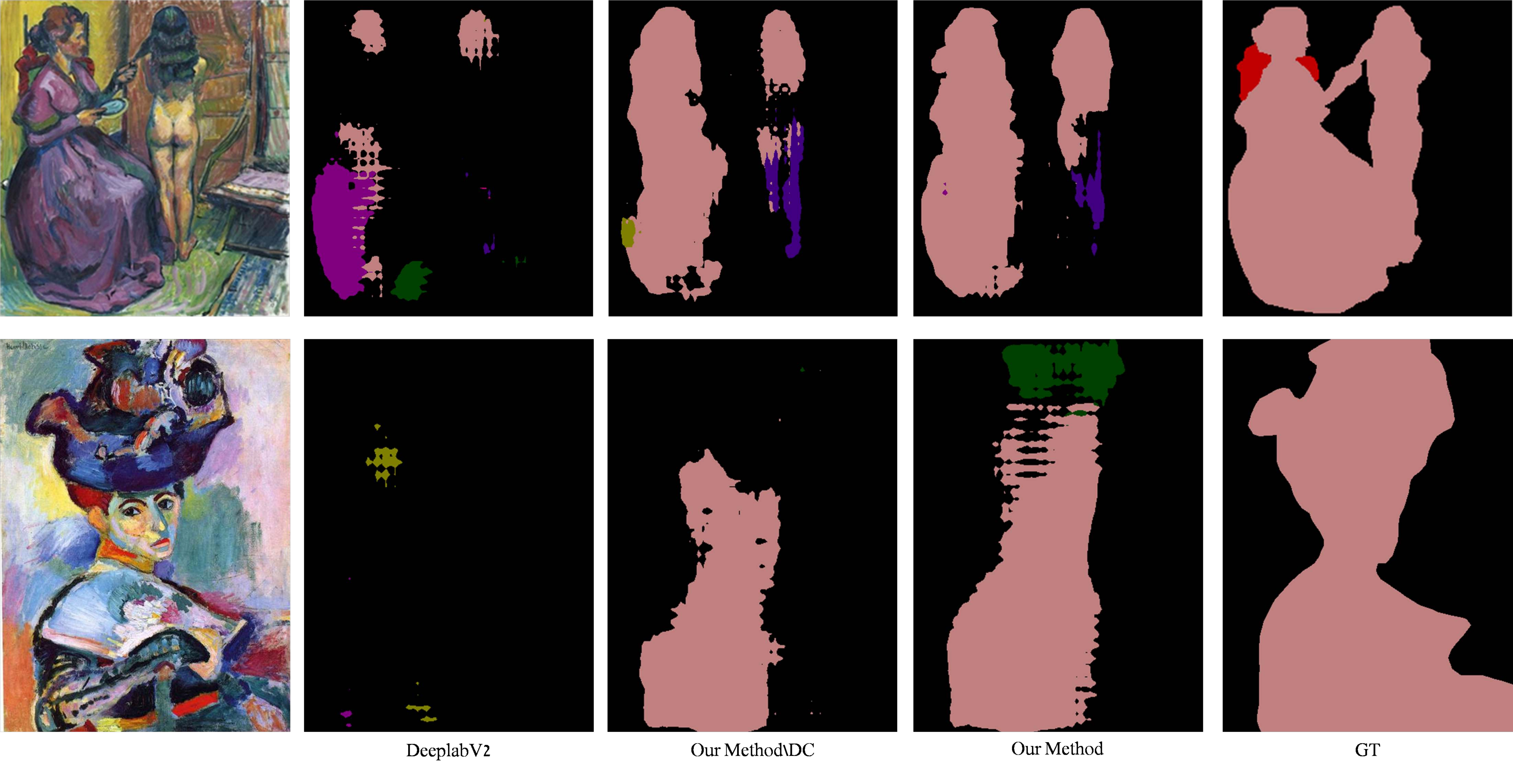}
 \hfill
    \caption{ Qualitative segmentation results from DRAM Fauvism unseen test set.}
    \label{fig:qualitative_fauvism}
\end{center}
\end{figure*}

\begin{figure*}[t]
\begin{center}
\includegraphics[width=1.0\textwidth]{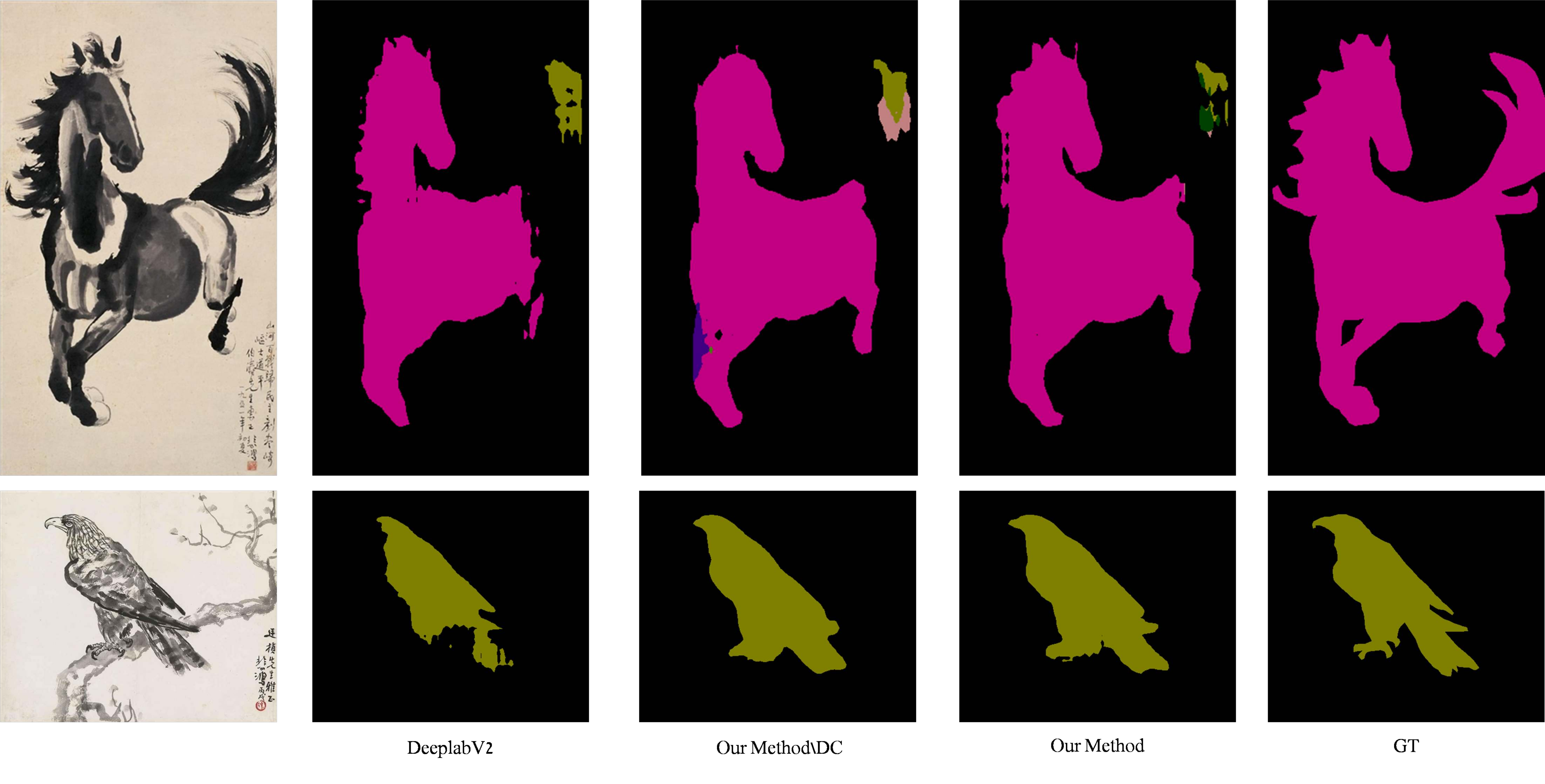}
 \hfill
    \caption{ Qualitative segmentation results from DRAM Ink \& Wash unseen test set.}
    \label{fig:qualitative_ink_and_wash}
\end{center}
\end{figure*}

\begin{figure*}[t]
\begin{center}
\includegraphics[width=1.0\textwidth]{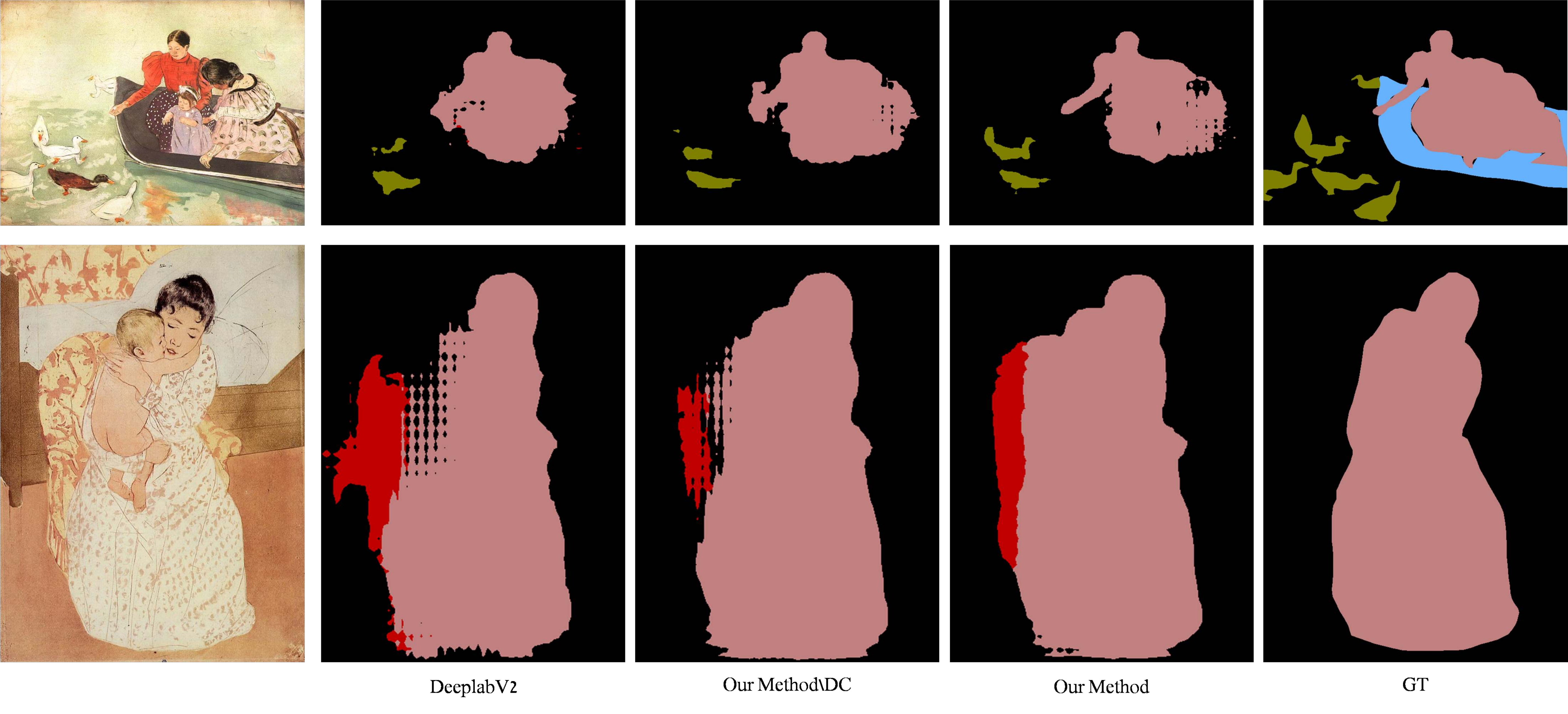}
 \hfill
    \caption{ Qualitative segmentation results from DRAM Japonism unseen test set.}
    \label{fig:qualitative_japonism}
\end{center}
\end{figure*}

\begin{figure*}[t]
\begin{center}
\includegraphics[width=1.0\textwidth]{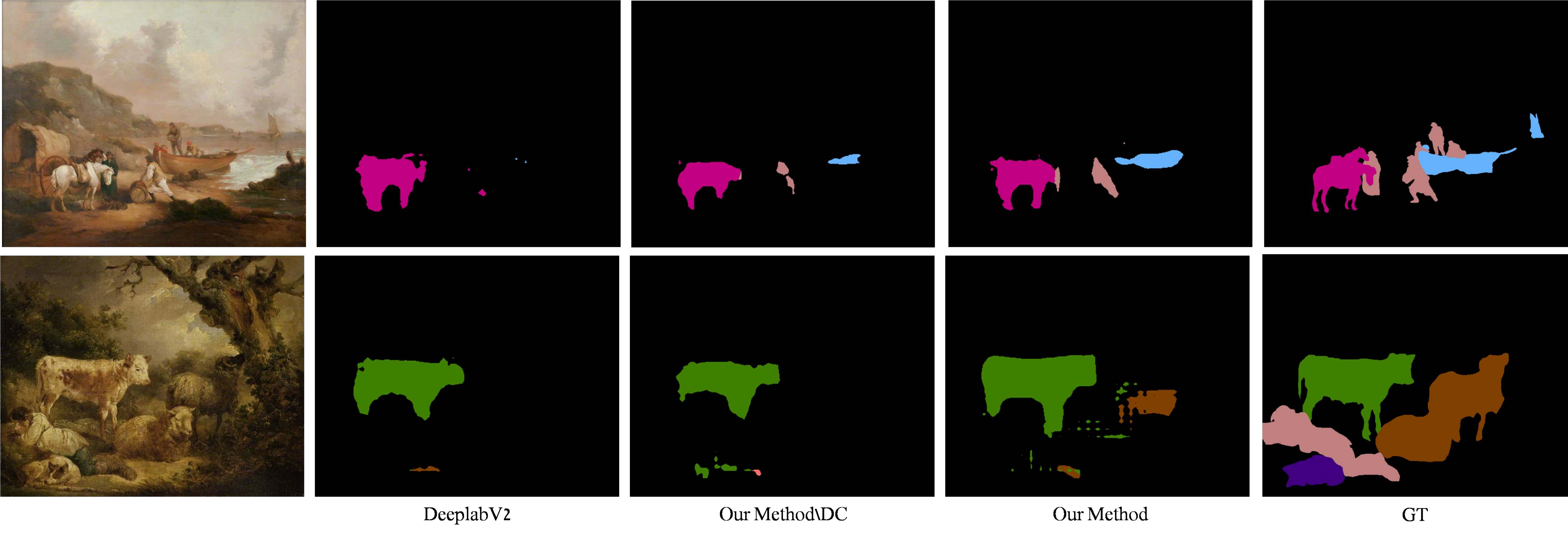}
 \hfill
    \caption{ Qualitative segmentation results from DRAM Rococo unseen test set.}
    \label{fig:qualitative_rococo}
\end{center}
\end{figure*}

{\ul Figure~\ref{fig:qualitative_realism}:} \\
A Summer Afternoon. Henry William Banks Davis. Wikiart. Public Domain. \\
His Only Friend. Briton-Riviere. Wikiart. Public Domain. \\
Sheep in Manger. Charles Jacque. Wikiart. Public Domain. \\
De Pianoles Door. Henriette Ronner Knip. Wikimedia. Public Domain. \\
Bouquet of flowers. Gustave Courbet. Wikiart. Public Domain. \\
\\
{\ul Figure~\ref{fig:qualitative_impressionism}:} \\
Children Playing with a Cat. Mary Cassatt. Wikiart. Public Domain. \\
A Black and White Cow.  Theodor Philipsen. Wikiart. Public Domain.\\
Sheperdess. Stefan Luchian. Wikiart. Public Domain.\\
Bathers at La Grenouillere. Claude Monet. Wikiart. Public Domain.\\
Head of a Dog. Pierre Auguste Renoir. Wikiart. Public Domain.\\
\\
{\ul Figure~\ref{fig:qualitative_post_impressionism}:} \\
The Cat. Gwen John. Wikiart. Public Domain.\\
Self-Portrait with Bandaged Ear. Vincent Van-Gogh. Wikiart. Public Domain.\\
The Jockey. Henri De-Toulouse Lautrec. Wikiart. Public Domain.\\
Girl Playing with a Dog (Vivette Terrasse). Pierre Bonnard. Wikiart. Public Domain.\\
Hangover. Henri De-Toulouse Lautrec Wikiart. Public Domain.\\
\\
{\ul Figure~\ref{fig:qualitative_expressionism}:} \\
Three Women at the Table by the Lamp. August Macke. Wikiart. Public Domain. \\
Roter Hund. Franz Marc. Wikimedia. Public Domain.\\
Death in the Sickroom. Edvard Munch. Wikiart. Public Domain.\\
Grazing Horses IV (The Red Horses). Franz Marc. Wikiart. Public Domain.\\
Cows at Sunset. Ernst Ludwig Kirchner. Wikiart. Public Domain.\\
\\
{\ul Figure~\ref{fig:qualitative_art_noveau}:} \\
Horsemen Riding in the Bois de Boulogne. Henri De-Toulouse Lautrec. Wikiart. Public Domain. \\
Cats. Theophile-Steinlen. Wikiart. Public Domain.\\
\\
{\ul Figure~\ref{fig:qualitative_baroque}:} \\
A Sleeping Dog with Terracotta Pot. Gerrit Dou. Wikiart. Public Domain.\\
Saint Thomas of Villanueva Dividing his Clothes Among Beggar Boys.  Bartolome Esteban Murillo. Wikiart. Public Domain.\\
\\
{\ul Figure~\ref{fig:qualitative_cubism}:} \\
Long Yellow Horse. Franz Marc. Wikiart. Public Domain. \\
Still Life with Pears. Maurice De-Vlaminck. Public Domain. \\
\\
{\ul Figure~\ref{fig:qualitative_divisiosnim}:} \\
Rückkehr Zum Schafstall. Giovanni Segantini. Wikiart. Public Domain.\\
Portrait von Oskar Kurt. Cuno Amiet. Wikiart. Public Domain US.\\
\\
{\ul Figure~\ref{fig:qualitative_fauvism}:} \\
Child at the Morning Toilette. Cuno Amiet. Wikiart. Public Domain US.\\
Woman with a Hat. Henri-Matisse. Public Domain US.\\
\\
{\ul Figure~\ref{fig:qualitative_ink_and_wash}:} \\
Galloping Horse. Xu Beihong. Wikiart. Public Domain China.\\
Hawk. Xu Beihong. Wikiart. Public Domain China.\\
\\
{\ul Figure~\ref{fig:qualitative_japonism}:} \\
Feeding the Ducks. Mary Cassatt. Wikiart. Public Domain.\\
Nude Child. Mary Cassatt. Wikiart. Public Domain.\\
\\
{\ul Figure~\ref{fig:qualitative_rococo}:} \\
Smugglers on a Beach. George Morland. Wikiart. Public Domain.\\
Calf and Sheep. George Morland. Wikiart. Public Domain.\\

\end{document}